\documentclass{article}

\PassOptionsToPackage{numbers, square}{natbib}

\usepackage[preprint]{neurips_2025}

\usepackage{amsmath}
\usepackage{amsfonts}
\usepackage{algorithm}
\usepackage{algorithmic}
\usepackage{graphicx}
\usepackage{caption}
\usepackage{subcaption}
\usepackage{hyperref}
\usepackage{multirow}

\usepackage{array}
\usepackage[utf8]{inputenc} 
\usepackage[T1]{fontenc}    
\usepackage{hyperref}       
\usepackage{url}            
\usepackage{booktabs}       
\usepackage{amsfonts}       
\usepackage{nicefrac}       
\usepackage{microtype}      
\usepackage{xcolor}         

\title{WATS: Calibrating Graph Neural Networks with Wavelet-Aware Temperature Scaling}

\author{%
  Xiaoyang Li \\
  Independent Researcher\\
  \texttt{Lxy289692485@gmail.com}\\
  \And
  Linwei Tao \\
  School of Computer Science \\
  University of Sydney \\
  \texttt{linwei.tao@sydney.edu.au} \\
  \AND
  Haohui Lu \\
  University of Sydney \\
  \texttt{haohui.lu@sydney.edu.au} \\
  \And
  Minjing Dong \\
  City University of Hong Kong\\
  \texttt{minjdong@cityu.edu.hk} \\
  \AND
  Junbin Gao \\
  University of Sydney \\
  \texttt{junbin.gao@sydney.edu.au} \\
  \And
  Chang Xu \\
  School of Computer Science \\
  University of Sydney \\
  \texttt{c.xu@sydney.edu.au} 
}

\begin{document}
\maketitle

\begin{abstract}
Graph Neural Networks (GNNs) have demonstrated strong predictive performance on relational data; however, their confidence estimates often misalign with actual predictive correctness, posing significant limitations for deployment in safety-critical settings. While existing graph-aware calibration methods seek to mitigate this limitation, they primarily depend on coarse one-hop statistics, such as neighbor-predicted confidence, or latent node embeddings, thereby neglecting the fine-grained structural heterogeneity inherent in graph topology. In this work, we propose Wavelet-Aware Temperature Scaling (WATS), a post-hoc calibration framework that assigns node-specific temperatures based on tunable heat-kernel graph wavelet features. Specifically, WATS harnesses the scalability and topology sensitivity of graph wavelets to refine confidence estimates, all without necessitating model retraining or access to neighboring logits or predictions. Extensive evaluations across seven benchmark datasets with varying graph structures and two GNN backbones demonstrate that WATS achieves the lowest Expected Calibration Error (ECE) among all compared methods, outperforming both classical and graph-specific baselines by up to 42.3\% in ECE and reducing calibration variance by 17.24\% on average compared with graph-specific methods. Moreover, WATS remains computationally efficient, scaling well across graphs of diverse sizes and densities. Code will be released based on publication.
\end{abstract}

\section{Introduction}

GNNs offer a principled approach for learning over structured data and have achieved strong empirical performance across a wide range of domains, including social network modeling \citep{fan2019graphneuralnetworkssocial}, traffic forecasting \citep{su151511893}, and healthcare applications \citep{10115472, lu2021weighted}.  
They support key tasks such as node classification \citep{10.1145/3437963.3441720, Sun_Zhang_Mou_Zhu_Xiong_Zhang_2022}, link prediction \citep{NEURIPS2018_53f0d7c5, NEURIPS2023_fba4a59c}, and graph-level inference \citep{NEURIPS2021_85267d34, DBLP:journals/corr/abs-2106-07971}. While GNNs are widely adopted for their representational power, their output confidence often fails to reflect true predictive reliability, which is an issue of growing concern in high-stakes domains, for instance, medical diagnosis and financial risk assessment.  

Model calibration, which measures the alignment between a model's predicted confidence and its true correctness likelihood \citep{DBLP:journals/corr/GuoPSW17}, is a key aspect of model reliability.  
A well-calibrated model is expected to produce predictions whose confidence scores accurately reflect the observed accuracy. For example, a prediction made with 70\% confidence should be correct 70\% of the time.
Recent findings highlight that GNNs behave differently from standard independent and identically distributed ($i.i.d.$) trained models such as CNNs and transformers \citep{9775118, Tao_Dong_Xu_2025, wen2024mitigating}.  
Unlike CNNs and transformers, which often suffer from overconfidence, GNNs tend to be systematically underconfident: their predicted confidence scores are consistently lower than their true accuracy \citep{wang2022gcl,hsu2022makes,liu2022calibration}.  

Several recent approaches aim to address this issue through graph-aware calibration strategies, including Graph Attention Temperature Scaling (GATS) \citep{hsu2022makes}, CaGCN \citep{wang2021confident}, Graph Ensemble Temperature Scaling (GETS) \citep{zhuang2024gets}, and SimCalib \citep{tang2024simcalib}. These approaches typically enhance node-level calibration by integrating neighbour structural cues with their predictive status, such as confidence level. 

However, they predominantly rely on shallow neighborhood statistics or opaque latent representations, which may result in unstable and inaccurate uncertainty estimation. As they capture only limited, local information and fail to reflect the broader structural context, leading to unreliable calibration, especially for low-degree scenarios. For illustration, we provide empirical evidence that nodes with similar local statistics can exhibit different miscalibration levels across graphs.
Based on the above limitation, we aim to build a method that:
(i)  flexibly incorporates neighborhood information without relying on additional explicit pretraining status
(ii) maintains high calibration performance across diverse graph domains, while remaining lightweight and post-hoc.
(iii) performs temperature scaling at the node level to allow identical correction based on multi-hop structural information.

Therefore, in this work, we propose a calibration framework called \textsc{Wavelet-Aware Temperature Scaling} (\textsc{WATS}), which introduces flexibly scaled structural features through graph wavelets.
We incorporate graph wavelets because they offer a principled way to capture structural information at multiple scales \citep{donnat2018learning}. Our methods, wavelets spatial localization controlled by the scale parameters $s$ and $k$, enabling fine-grained capture of multi-hop dependencies. Differs from graph wavelet algorithms used in neural network \citep{donnat2018learning, behmanesh2022geometric} in that it does not aim to reconstruct or smooth node features for node classification. Instead, it uses wavelet coefficients as structural signatures to indicate the node uncertainty, which allows WATS to adaptively calibrate predictions based on the underlying structure, helping to correct confidence errors where standard methods fall short.
Our main contributions are summarized as follows:

\textbf{WATS}: We propose a novel post-hoc calibration framework that performs node-wise temperature scaling using flexibly scaled graph wavelet features.
    
\textbf{Robust, interpretable structural features: }We show that graph wavelet features are stable and geometry-aware, encoding local graph structure without relying on potentially noisy signals such as neighboring logits or distances to labeled nodes. 
    
\textbf{Extensive empirical validation:} Across multiple graph benchmarks and GNN architectures. \textsc{WATS} consistently improves calibration quality, reducing ECE and outperforming both classical and graph-specific calibration baselines.

\section{Related Work}

\subsection{Uncertainty Calibration for Neural Networks}
Uncertainty calibration aims to align a model’s predicted confidence with the true likelihood of correctness. Existing approaches are typically divided into two categories: in-training and post-hoc

\textbf{In-training} approaches incorporate uncertainty estimation within the model optimization process. For example, Bayesian Neural Networks (BNNs) and variational inference methods achieve this by imposing probabilistic distribution over model parameters  \citep{gal2016dropout, MACKAY199573, springenberg2016bayesian}. Alternative frequentist strategies, such as conformal prediction \citep{tibshirani2019conformal} and quantile regression \citep{romano2019conformalized}, are also employed to generate calibrated probability estimates.

\textbf{Post-hoc} methods, in contrast, calibrate a pre-trained model without modifying its internal parameters. These include non-parametric techniques such as histogram binning \citep{zadrozny2001learning} and isotonic regression \citep{zadrozny2002transforming}, as well as parametric approaches that assume a specific transformation form, including temperature scaling (TS) \citep{DBLP:journals/corr/GuoPSW17} and Beta Calibration \citep{kull2017beta}), to adjust the model’s output logits accordingly.

\subsection{Graph-Specific Calibration Methods}
While post-hoc calibration methods perform well on Euclidean data with CNNs, their effectiveness declines on graph-structured data due to the lack of relational modeling \citep{hsu2022makes, wang2021confident}. To address this, several graph-aware approaches have been proposed. CaGCN \citep{wang2021confident} uses a GCN-based temperature predictor to incorporate structure, while GATS \citep{hsu2022makes} applies attention over neighborhoods for node-specific temperatures. GETS \citep{zhuang2024gets} introduces a sparse mixture-of-experts using degree, features, and confidence. SimCalib \citep{tang2024simcalib} adds similarity-preserving regularization, and \citet{shi2023calibrate} use reinforcement learning to adapt calibration to graph structure. Beyond post-hoc methods, \citet{yang2024calibrating} reweight edges during training to improve calibration, and \citet{yang2024balanced} propose a calibration-aware loss targeting underconfidence caused by shallow GNNs. These methods collectively integrate graph structure and node-level signals to enhance calibration.

\subsection{Graph Wavelet}

Graph wavelets provide compact, spatially localized bases that are well suited to graph signal processing and structural representation learning.  Whereas classical wavelets such as Haar and Daubechies are defined on Euclidean lattices \citep{540087,RESNIKOFF1992273}, graph wavelets extend these ideas to non-Euclidean domains by leveraging the spectral properties of the graph Laplacian \citep{das2004laplacian,shuman2013emerging,xu2019graph}.  

A particularly versatile construction is the lifting scheme \citep{sweldens1998lifting}, which can be transferred to graphs without any data-driven training.  \citet{hammond2011wavelets} further improved practicality by replacing the costly Laplacian eigendecomposition with Chebyshev polynomial approximations, enabling efficient wavelet transforms on large graphs.  Since then, graph wavelets have supported a variety of downstream tasks, including graph convolutional architectures \citep{xu2019graph, Deb_Rahman_Rahman_2024}, multimodal wavelet networks \citep{behmanesh2022geometric}, community detection via scale-adaptive filtering \citep{tremblay2014graph}, and diffusion-based node embeddings that capture multi-scale structural patterns \citep{donnat2018learning}. All these methods and application indicates the importance of Graph wavelet in both theoretical and empirically practice

\section{Method}

\subsection{Preliminary Study}

We address the problem of uncertainty calibration in semi-supervised node classification tasks over graphs.  
Let \(\mathcal{G} = (\mathcal{V}, \mathcal{E})\) denote a graph, where \(\mathcal{V}\) is the set of nodes and \(\mathcal{E}\) is the set of edges.  
The adjacency matrix is denoted by \(A \in \mathbb{R}^{N \times N}\), where \(N = |\mathcal{V}|\).  
Each node \(v_i\in\mathcal{V}\) has a feature \(x_i\in\mathcal{X}\), and for a subset of labeled nodes \(\mathcal{L}\subseteq\mathcal{V}\), the true label \(y_i\in\{1,\ldots,K\}\) is provided.  
Let \(X=[x_1,\ldots,x_N]^\top\) be the feature matrix and \(Y=[y_1,\ldots,y_N]^\top\) the label vector,
a GNN \(f_\theta\) performs node classification via predicting node-wise class probabilities \(\hat p_i(y)\) as
\[
\hat y_i = \arg\max_y \hat p_i(y), 
\quad 
\hat c_i = \max_y \hat p_i(y),
\]
where $y_i$ denotes the predicted label and $c_i$ denotes its confidence. In the field of model calibration, a well-calibrated model provides confidence that aligns with the true accuracy well as
\[
\mathbb{P}(y_i = \hat y_i \mid \hat c_i = c) = c
\quad\forall\,c\in[0,1].
\]
The measurement of model calibration can be computed via Expected Calibration Error (ECE) \citep{DBLP:journals/corr/GuoPSW17} as $\mathbb{E}[|\mathbb{P}(y_i = \hat y_i | \hat c_i)-c]$, however, ECE cannot be easily computed due to limited samples. Thus, an estimation of ECE is introduced by grouping samples into \(M\) bins with equal confidence intervals as $B_m = \bigl\{\,j\in\mathcal{N}\mid \tfrac{m-1}{M} < \hat c_j \le \tfrac{m}{M}\bigr\}$, where \(\mathcal{N}\) is the subset of node that used in evaluation \(\mathcal{N}\subseteq\mathcal{V}\). Given the bin accuracy $\mathrm{Acc}(B_m)
= \frac{1}{|B_m|}\sum_{i\in B_m}\mathbf{1}(\hat y_i=y_i)$ and bin confidence $\mathrm{Conf}(B_m)
= \frac{1}{|B_m|}\sum_{i\in B_m}\hat c_i$, the approximation can be achieved by computing the expected difference between bin accuracy and confidence as
\begin{equation}
\scalebox{1.0}{$ \displaystyle
\mathrm{ECE} = \sum_{m=1}^M \frac{|B_m|}{|\mathcal{N}|}\,\bigl|\mathrm{Acc}(B_m)-\mathrm{Conf}(B_m)\bigr|.
$}
\end{equation}


\subsection{Uncertainty Estimation in GNNs}
\paragraph{Calibrate via One-hop Statistics} 
Message-passing is widely adopted in GNNS, including  GCN \citep{kipf2016semi} and GAT \citep{velivckovic2017graph}, which can be simplified via a degree-normalized mean aggregator as
\begin{equation}
h_i^{(\ell+1)}
= \frac{1}{d_i + 1}\Bigl(h_i^{(\ell)} + \sum_{j\in\mathcal N(i)}h_j^{(\ell)}\Bigr),
\quad
d_i = |\mathcal N(i)|.
\end{equation}
where \(d_i\) excludes the node itself, so \(d_i+1\) accounts for the self-loop.  The final embedding \(h_i^{(L)}\) induces the confidence \(\hat c_i\).  
This local aggregation implicitly determines the final prediction and the associated confidence $\hat{c}_i$ of node $i$. 
Although GATS confines all structural operations, including neighbor temperature aggregation, attention weights, neighbor confidence averaging to 1-hop, and CaGCN and GETS stack two GCN layers to nominally reach 2-hop, each layer itself still performs only 1-hop aggregation. As a result, these methods are unable to adaptively capture longer-range dependencies.
Although these calibration techniques show effectiveness, we argue that one-hop statistics only cannot provide an accurate estimation of node uncertainty in GNNs. Considering a simplified one-hop estimator of confidence:
\[
\hat c_i \approx \frac{1}{d_i+1}\sum_{j\in\{i\}\cup\mathcal N(i)} y_j,
\]
where \(y_j\in\{0,1\}\) is the true label indicator for node \(j\).  Then the per-node calibration bias is
\begin{equation} \label{eq:node_bias}
\mathrm{bias}_i
= \bigl|\hat c_i - \mathbf{1}(\hat y_i=y_i)\bigr|
\approx
\Bigl|\,y_i - \frac{1}{d_i+1}\sum_{j\in\mathcal N(i)}y_j\Bigr|.
\end{equation}  

\begin{figure}[!ht]
\vspace{-5pt}
     \centering
     \begin{subfigure}[b]{0.45\textwidth}
         \centering
         \includegraphics[width=\textwidth]{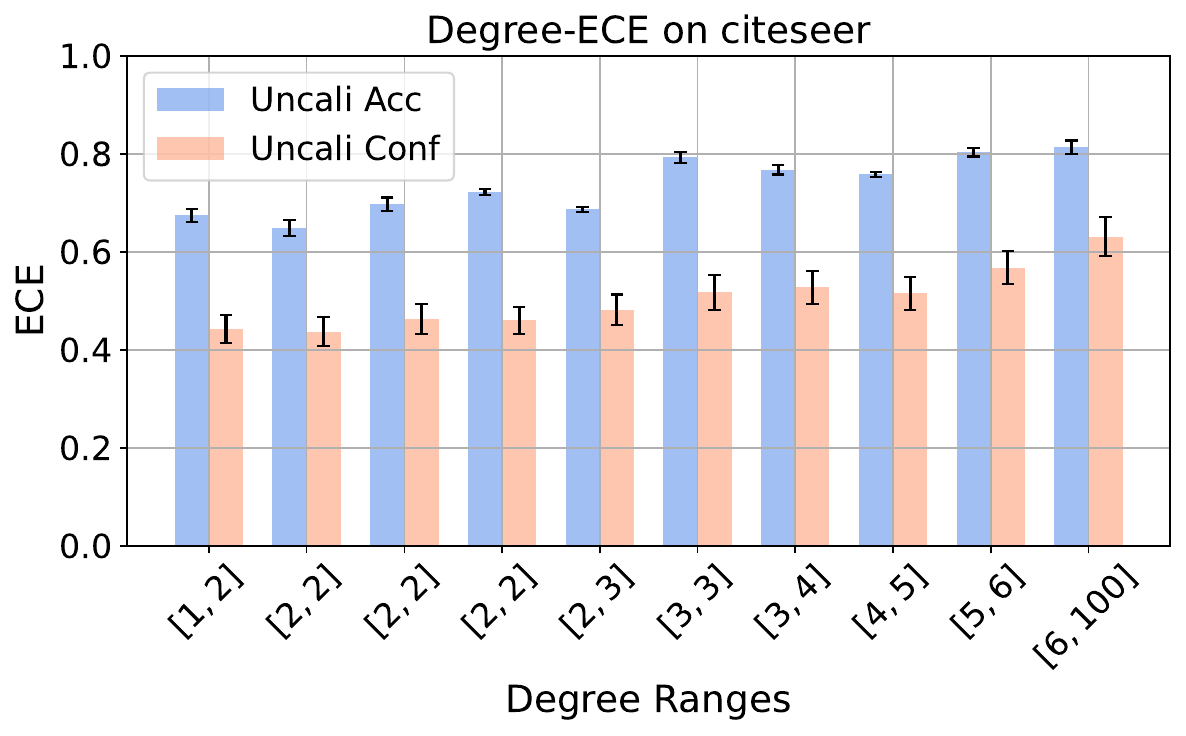}
         \label{fig:cora}
     \end{subfigure}
    \hfill
     \begin{subfigure}[b]{0.45\textwidth}
         \centering
         \includegraphics[width=\textwidth]
         {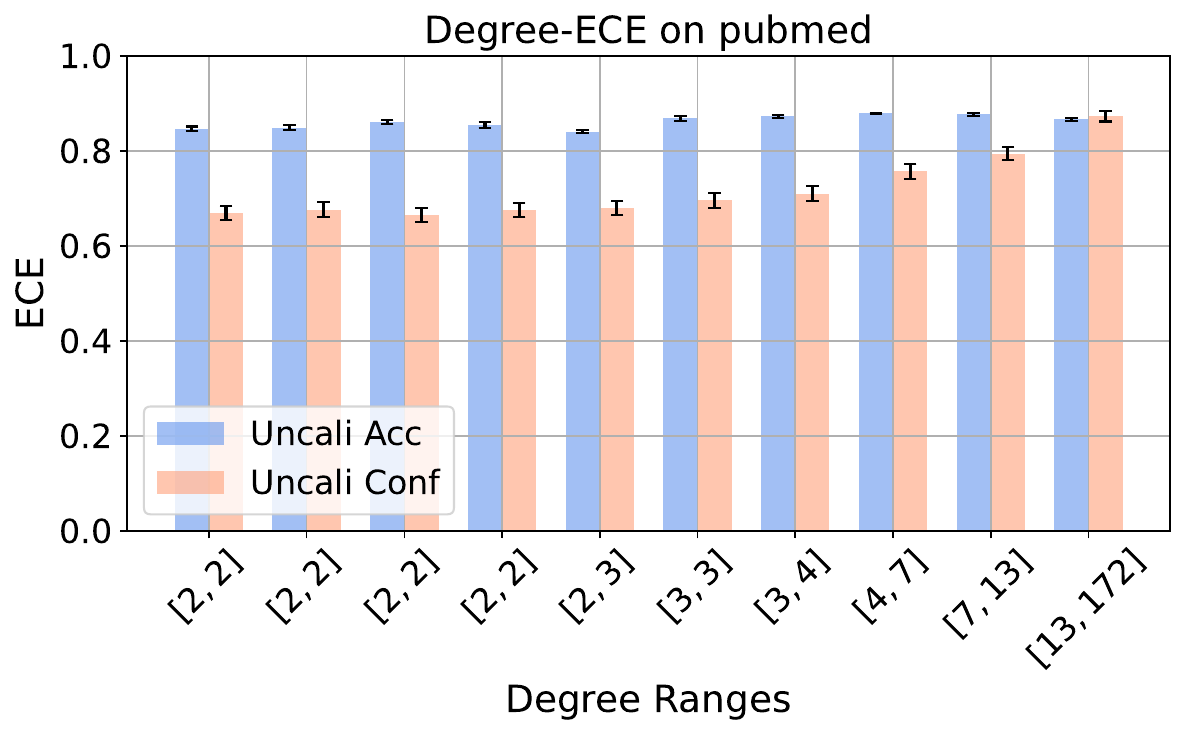}
         \label{fig:pubmed}
     \end{subfigure}
     \caption{The x-axis shows node degree ranges and the y-axis represents the ECE. The blue bars indicate the uncalibrated accuracy and the pink bars indicate the uncalibrated confidence. Compared to Pubmed, Citeseer exhibits a more noticeable gap between accuracy and confidence in low-degree ranges, highlighting stronger underconfidence in sparse regions.}

     \label{fig:2 graphs deg}

\end{figure}

As shown in Eq. \ref{eq:node_bias}, for example, when $d_i = 2$ and the neighbor labels are $[0,1]$, the average is $1/3$ regardless of the true label $y_i$, making the estimate uninformative, which means high uncertainty. In sparse or low-homophily regions, one-hop neighborhoods may carry weak or misleading signals, resulting in poor bias approximation. This motivates the need for structure-aware calibration methods that incorporate richer, multi-hop graph information.
This expression highlights the critical influence of node degree on predictive uncertainty. Nodes with low degree have limited access to neighborhood information, which in GNNs systematically manifests as underconfidence. That compile with the previous study \citep{hsu2022makes, wang2021confident} and illustrated in Figure~\ref{fig:2 graphs deg}. Conversely, as degree increases, the richer information integration drives the model toward higher confidence estimates.

Moreover, as shown in Figure~\ref{fig:2 graphs deg}, we observe that even within the same degree range, the ECE varies significantly across graphs. While the overall trend that higher-degree nodes tend to have higher confidence is consistent, the magnitude and pattern of this trend differ per graph. This discrepancy highlights that node degree alone is insufficient to explain calibration behaviors in GNNs.

\paragraph {Calibrate via Local Structure}

While node degree is highly correlated with predictive confidence, this relationship is not consistent across graphs. Our analysis reveals that nodes with similar degree ranges can exhibit markedly different levels of miscalibration across datasets, indicating that degree is informative but insufficient to fully explain confidence behavior in GNNs. Additionally, the confidence level can also be unstable, making it an unreliable indicator of confidence.

These observations point to a deeper insight: node confidence is influenced not only by immediate neighbors and structures, but also by the broader structural context in which a node is situated. This is further supported by \citet{wang2022gcl}, who demonstrate that although deeper GNNs may suffer from reduced accuracy, they tend to exhibit higher confidence, highlighting the importance of extended neighborhood influence. Prior works~\cite{wang2022gcl, hsu2022makes, tang2024simcalib} have emphasized some graph features, such as homophily and node similarity. GETS~\cite{zhuang2024gets} partially addresses this by incorporating degree embeddings into a mixture-of-experts framework. While all these methods lack the ability to capture multi-hop structural patterns or incorporate unstable confidence.
Therefore, addressing the above limitation and improving the calibration requires leveraging multi-hop context and structure-aware features, enabling the model to better handle both under- and over-confidence in structurally diverse regions of the graph.

\subsection{Wavelet-Aware Temperature Scaling}

We propose WATS, a lightweight and effective node-wise calibration framework that can be seamlessly applied to any pretrained GNN with scalability to large graphs. Unlike conventional or graph-specific post-hoc methods that rely on global or one-hop features, WATS introduces a structural perspective by leveraging graph wavelet features with tunable scales.

These wavelet representations capture rich, scalable structural signals~\citep{hammond2011wavelets, crovella2003graph}, often neglected in calibration. By learning a temperature for each node based on its structural embedding, WATS aligns confidence with correctness in a fine-grained, node-specific manner. In addition to its strong empirical performance, WATS is also architecture-agnostic, making it broadly applicable across diverse graph types and calibration scenarios.

\subsubsection{Graph Wavelet Transform}

Traditional graph signal processing often relies on graph Fourier transform, which projects signals into the spectral domain using the eigenvectors of the normalized graph Laplacian $\mathbf{L}_{\text{sym}} = \mathbf{I} - \mathbf{D}^{-1/2} \mathbf{A} \mathbf{D}^{-1/2}$ as orthonormal bases.
Given a signal $\mathbf{x} \in \mathbb{R}^N$, its Fourier transform is defined as $\hat{\mathbf{x}} = \mathbf{U}^\top \mathbf{x}$ and the inverse as $\mathbf{x} = \mathbf{U} \hat{\mathbf{x}}$, where $\mathbf{U}$ contains the eigenvectors of $\mathbf{L}_{\text{sym}}$ \citep{shuman2013emerging}.
While this formulation enables spectral filtering via $\mathbf{U} g_\theta \mathbf{U}^\top \mathbf{x}$, it suffers from several limitations \citep{hammond2011wavelets, xu2019graph, zheng2021framelets}{:
(1) The eigendecomposition of $\mathbf{L}_{\text{sym}}$ has high computational cost ($\mathcal{O}(N^3)$);
(2) $\mathbf{U}$ is generally dense, making the transform costly for large graphs;
(3) The resulting filters lack localization in the vertex domain, limiting their ability to capture localized structural patterns.}

To overcome these issues, we adopt the graph wavelet transform, which retains the spectral benefits of Fourier analysis while introducing localization and sparsity.
Graph wavelet bases are constructed using a heat kernel scaling function $g(s\lambda) = e^{-s\lambda}$, where $s > 0$ is a scale parameter controlling the diffusion extent.
The wavelet operator is defined as:
\begin{equation}
\boldsymbol{\Psi}_s = \mathbf{U} \, \text{diag}(g(s\lambda_1), \dots, g(s\lambda_N)) \, \mathbf{U}^\top
\end{equation}
where $\lambda_i$ are the eigenvalues of $\mathbf{L}_{\text{sym}}$.
The inverse transform uses $g(-s\lambda)$, yielding efficient localized filtering analogous to diffusion. 
Direct computation of $\boldsymbol{\Psi}_s$ is still impractical for large graphs.
To address this, we adopt the Chebyshev polynomial approximation to avoid explicit eigendecomposition, following  \citep{hammond2011wavelets, xu2019graph}.
We first rescale $\mathbf{L}_{\text{sym}}$ as:
$
\hat{\mathbf{L}} = \frac{2}{\lambda_{\max}} \mathbf{L}_{\text{sym}} - \mathbf{I}, \quad \lambda_{\max} \approx 2$
and define Chebyshev polynomials $\{\mathbf{T}_k\}_{k=0}^K$ via the recurrence:
$\mathbf{T}_0 = \mathbf{X}_0, 
\mathbf{T}_1 = \hat{\mathbf{L}} \mathbf{X}_0, 
\mathbf{T}_k = 2 \hat{\mathbf{L}} \mathbf{T}_{k-1} - \mathbf{T}_{k-2}, \quad \text{for } k \geq 2$ where $\mathbf{X}_0$ is the initial input signal.
In our setting, we choose $\mathbf{X}_0$ as the $\log$-degree to preserve structural properties while mitigating skewed degree distributions. Degree encodes a node's connectivity and its potential for information aggregation in message-passing GNNs, and in previous section, it is proven to be an essential factor of uncertainty. 
The final wavelet-transformed feature matrix is constructed as:
\begin{equation}
\mathbf{S} = \sum_{k=0}^K \alpha_k \mathbf{T}_k, \quad \text{with } \alpha_k = e^{-s k}
\end{equation}  Here we are using $K$-order Chebyshev polynomial to approximate the wavelet scaling function $g(s\lambda) = e^{-s\lambda}$, that is 
\[
g(s\lambda) \approx \frac12 c_0 + \sum^K_{k=1}c_k T_k(\lambda)
\]
with 
\[
c_k = \frac2{\pi} \int^1_{-1} \frac{T_k(y)g(sy)}{\sqrt{1-y^2}} dy
\]
refer to \citet{hammond2011wavelets} $c_k$ are computable constants before training. Followed by row-wise $\ell_1$ normalization:
\begin{equation}
\mathbf{H}_i = \frac{\mathbf{S}_i}{\|\mathbf{S}_i\|_1}, \quad \forall i \in \{1, \dots, N\}
\end{equation}
The hyper-parameter \(k\) sets the maximum receptive-field size (i.e., the number of hops considered), while the scale parameter \(s\) governs the extent of diffusion. A small \(s\) restricts diffusion and thus accentuates local structure, whereas a large \(s\) allows more extensive diffusion, leading to stronger smoothing and the integration of broader, long-range context. In practice, selecting appropriate values for \(k\) and \(s\) enables control over the locality and granularity of the wavelet features. This flexibility is crucial for capturing diverse structural patterns across graphs of varying density and topology.

\subsubsection{Node-wise Temperature Scaling}

Based on the extracted wavelet features, we predict a node-specific temperature parameter to rescale the logits produced by the original GNN.  
Given the feature matrix $\mathbf{H} \in \mathbb{R}^{N \times (K+1)}$, we employ a two-layer multilayer perceptron (MLP) to capture the non-linear relationship and predict the temperatures:
\begin{equation}
\tau_i = \text{Softplus}(\text{MLP}(\mathbf{h}_i))
\end{equation}
where $\mathbf{h}_i$ is the wavelet feature vector for node $i$, and Softplus ensures the positivity of the predicted temperatures.  
This design provides a flexible and efficient mechanism for uncertainty calibration across the graph.
The calibrated logits are obtained via post-hoc temperature scaling:
\[
\tilde{z}_i = \frac{z_i}{\tau_i}
\]
where $z_i$ is the original output logit from the GNN, and \(\tilde{z}_i\) is the rescaled logit after calibration. The temperature predictor is trained by minimizing the cross-entropy loss on the validation set using the rescaled logits.

\section{Experiment}
\subsection{Experiment Setting}

We evaluate the calibration performance of our proposed WATS method on seven widely-used graph datasets: Cora \citep{mccallum2000automating}, Citeseer \citep{giles1998citeseer}, Pubmed  \citep{sen2008collective}, Cora-Full \citep{bojchevski2017deep}, Computers \citep{shchur2018pitfalls}, Photo \citep{shchur2018pitfalls}, and Reddit \citep{hamilton2017inductive}.  
These datasets cover a range of graph sizes, feature dimensions, and label complexities, providing a comprehensive benchmark for calibration analysis, detailed graph summary is shown on Appendix.

Following previous practice \citep{wang2021confident, hsu2022makes, tang2024simcalib}, we adopt two commonly used GNN architectures as base models, which are GCN \citep{kipf2016semi} and GAT \citep{velivckovic2017graph}. The models are trained under a semi-supervised node classification setting. After training, we perform post-hoc calibration using different methods without modifying the model parameters.

\paragraph{Training Settings.} Follow the experiment settings \citep{hsu2022makes, tang2024simcalib, zhuang2024gets}, We randomly use 20\% of nodes for training, 10\% for validation and calibration training, and 70\% for testing. The detail of training setting is given in the Appendix.
\paragraph{Calibration Settings.} For each method, calibration parameters are learned on the validation set and evaluated on the test set. Calibration performance is measured using the ECE with 10 bins. The detailed calibration setting are displayed in detail in Appendix.

\paragraph{Calibration Methods Compared.}
We compare several post-hoc calibration methods. TS applies a global temperature to all logits \citep{DBLP:journals/corr/GuoPSW17}, while ETS averages predictions from multiple temperature-tuned models \citep{zhang2020mix}. CaGCN uses a lightweight GCN to learn node-specific temperatures \citep{wang2021confident}, and GATS employs attention-based aggregation over one-hop neighbors \citep{hsu2022makes}. GETS introduces a sparse mixture-of-experts that combines degree, features, and logits \citep{zhuang2024gets}. WATS, our proposed method, predicts temperatures using tunable graph wavelet features and rescale logits.

\begin{table}[h!]
\scriptsize
\caption{Each result is reported as the mean ± standard deviation over 10 runs. ‘Uncalib’ refers to uncalibrated outputs, and ‘oom’ indicates out-of-memory failures where the method could not complete. Best performance on ECE are highlighted for each configuration.}
\label{tab:calibration-results}
\centering
\begin{tabular}{llccccccc}
\toprule
\textbf{Dataset} & \textbf{Model} & \textbf{Uncalib} & \textbf{TS} & \textbf{ETS} & \textbf{CAGCN} & \textbf{GATS} & \textbf{GETS} & \textbf{WATS} \\
\midrule
\multirow{2}{*}{Citeseer} & GCN & 23.20 $\pm$ 3.21 & 2.57 $\pm$ 0.78 & 3.45 $\pm$ 1.03 & 4.44 $\pm$ 1.47 & 2.38 $\pm$ 0.65 & 4.09 $\pm$ 1.36 & \textbf{2.15 $\pm$ 0.48 } \\
 & GAT & 15.61 $\pm$ 1.14 & 3.22 $\pm$ 0.29 & 3.55 $\pm$ 0.41 & 2.77 $\pm$ 0.59 & 3.22 $\pm$ 0.24 & 3.80 $\pm$ 2.05 & \textbf{2.67 $\pm$ 0.38} \\
\midrule
\multirow{2}{*}{Computers} & GCN & 5.94 $\pm$ 0.52 & 3.88 $\pm$ 0.70 & 3.91 $\pm$ 0.49 & 2.04 $\pm$ 0.34 & 3.34 $\pm$ 0.61 & 2.94 $\pm$ 1.26 & \textbf{1.31 $\pm$ 0.29} \\
 & GAT & 5.86 $\pm$ 1.26 & 2.12 $\pm$ 0.19 & 2.11 $\pm$ 0.20 & 2.99 $\pm$ 0.64 & 2.01 $\pm$ 0.17 & 3.95 $\pm$ 3.73 & \textbf{1.83 $\pm$ 0.28} \\
\midrule
\multirow{2}{*}{Cora} & GCN & 22.44 $\pm$ 1.17 & 2.25 $\pm$ 0.33 & 2.20 $\pm$ 0.44 & 2.79 $\pm$ 0.50 & 2.98 $\pm$ 0.59 & 2.96 $\pm$ 0.47 & \textbf{2.13 $\pm$ 0.51} \\
 & GAT & 17.26 $\pm$ 0.38 & 2.03 $\pm$ 0.31 & \textbf{1.92 $\pm$ 0.31} & 2.56 $\pm$ 0.38 & 2.15 $\pm$ 0.30 & 2.97 $\pm$ 0.47 & 2.02 $\pm$ 0.30 \\
\midrule
\multirow{2}{*}{Cora-full} & GCN & 27.79 $\pm$ 0.22 & 5.06 $\pm$ 0.10 & 5.00 $\pm$ 0.09 & 3.87 $\pm$ 0.22 & 5.13 $\pm$ 0.10 & 3.11 $\pm$ 1.95 & \textbf{2.04 $\pm$ 0.13} \\
 & GAT & 37.21 $\pm$ 0.37 & 2.50 $\pm$ 0.23 & 1.32 $\pm$ 0.16 & 4.79 $\pm$ 0.34 & 2.70 $\pm$ 0.26 & 2.16 $\pm$ 1.11 & \textbf{1.30 $\pm$ 0.23} \\
\midrule
\multirow{2}{*}{Photo} & GCN & 3.33 $\pm$ 0.22 & 2.45 $\pm$ 0.22 & 2.47 $\pm$ 0.20 & 1.72 $\pm$ 0.22 & 2.22 $\pm$ 0.19 & 3.25 $\pm$ 1.63 & \textbf{1.03 $\pm$ 0.18} \\
 & GAT & 3.21 $\pm$ 0.47 & 1.81 $\pm$ 0.43 & 2.34 $\pm$ 0.50 & 1.71 $\pm$ 0.10 & 1.80 $\pm$ 0.43 & 3.05 $\pm$ 1.67 & \textbf{1.63 $\pm$ 0.18} \\
\midrule
\multirow{2}{*}{Pubmed} & GCN & 14.33 $\pm$ 1.20 & 2.55 $\pm$ 0.38 & 2.81 $\pm$ 0.47 & 1.82 $\pm$ 0.36 & 2.30 $\pm$ 0.52 & 2.34 $\pm$ 0.51 & \textbf{1.04 $\pm$ 0.07} \\
 & GAT & 10.67 $\pm$ 0.30 & 0.88 $\pm$ 0.09 & \textbf{0.87 $\pm$ 0.09} & 0.91 $\pm$ 0.11 & 0.89 $\pm$ 0.08 & 0.90 $\pm$ 0.22 & \textbf{0.87 $\pm$ 0.08} \\
\midrule
\multirow{2}{*}{Reddit} & GCN & 6.69 $\pm$ 0.12 & 1.64 $\pm$ 0.05 & 1.64 $\pm$ 0.05 & 1.45 $\pm$ 0.08 & oom & 2.20 $\pm$ 0.36 & \textbf{1.02 $\pm$ 0.06} \\
 & GAT & 4.79 $\pm$ 0.16 & 3.29 $\pm$ 0.08 & 3.35 $\pm$ 0.12 & 0.73 $\pm$ 0.08 & oom & 1.10 $\pm$ 0.11 & \textbf{0.69 $\pm$ 0.13} \\
\bottomrule
\vspace{-10pt}
\end{tabular}
\end{table}

\subsection{Evaluation and Analysis}

We evaluate the calibration effectiveness of WATS across seven benchmark datasets and two representative GNN architectures (GCN and GAT), with results summarized in Table ~\ref{tab:calibration-results}. Empirical findings demonstrate that WATS consistently achieves the lowest ECE in most of scenarios, highlighting its efficacy in leveraging localized, flexibly scaled structural information for post-hoc uncertainty calibration. Beyond achieving superior average ECE scores, WATS also exhibits reduced standard deviations across runs, indicating improved robustness and stability compared to existing methods. These evidence prove that graph wavelet is able to capture sufficient local topology information to correct the confidence level. Moreover, even when the base model is already reasonably well calibrated, for example, on the Photo and Computers, WATS consistently delivers further reductions in calibration error, demonstrating its ability to adaptively refine predictive confidence across a range of baseline reliability levels.

\begin{figure}[h!]
    \centering
    \begin{subfigure}[b]{0.45\textwidth}
        \centering
        \includegraphics[width=\textwidth, height=4cm]{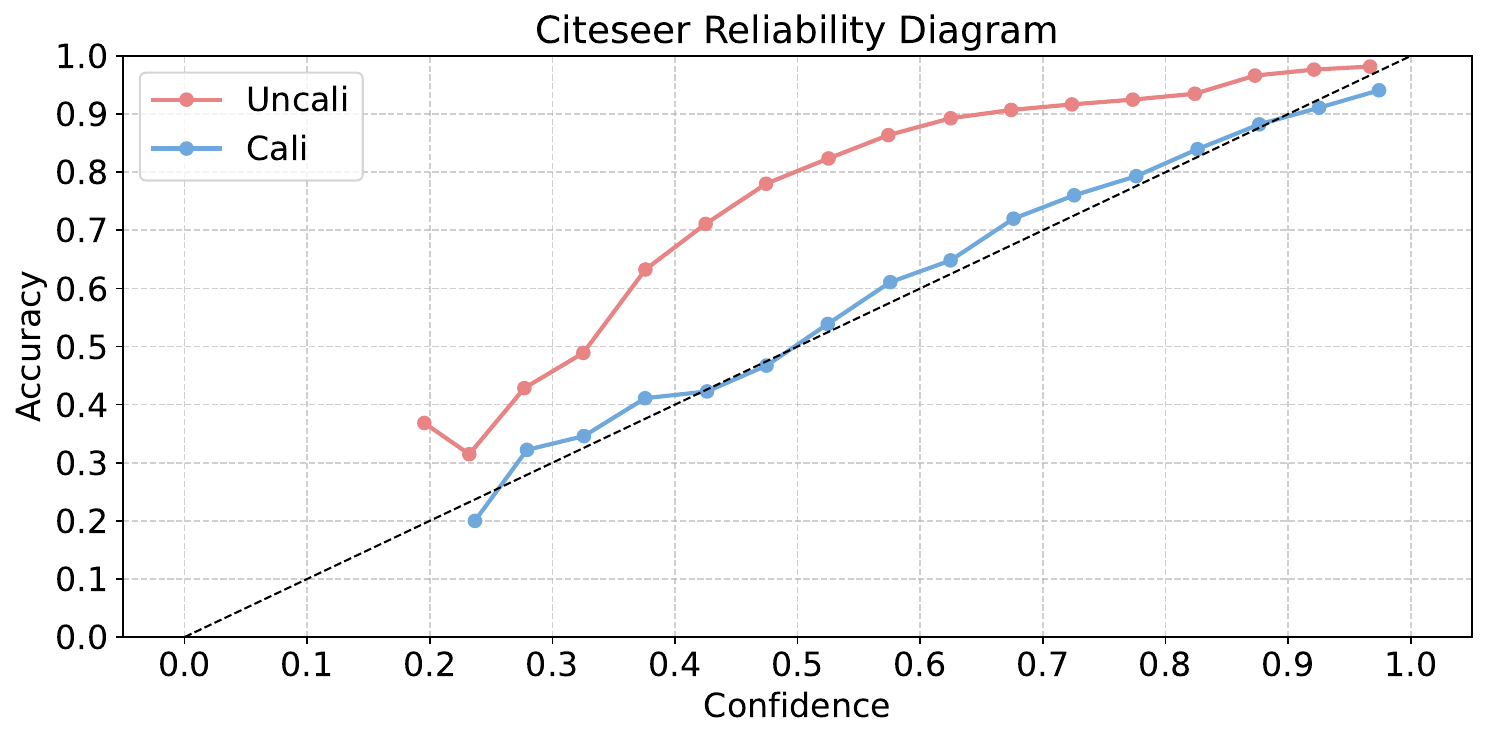}
        \caption{Reliability diagram on citeseer.}
        \label{fig:citeseer_reli}
    \end{subfigure}
    \hfill
    \begin{subfigure}[b]{0.45\textwidth}
        \centering
        \includegraphics[width=\textwidth, height=4cm]{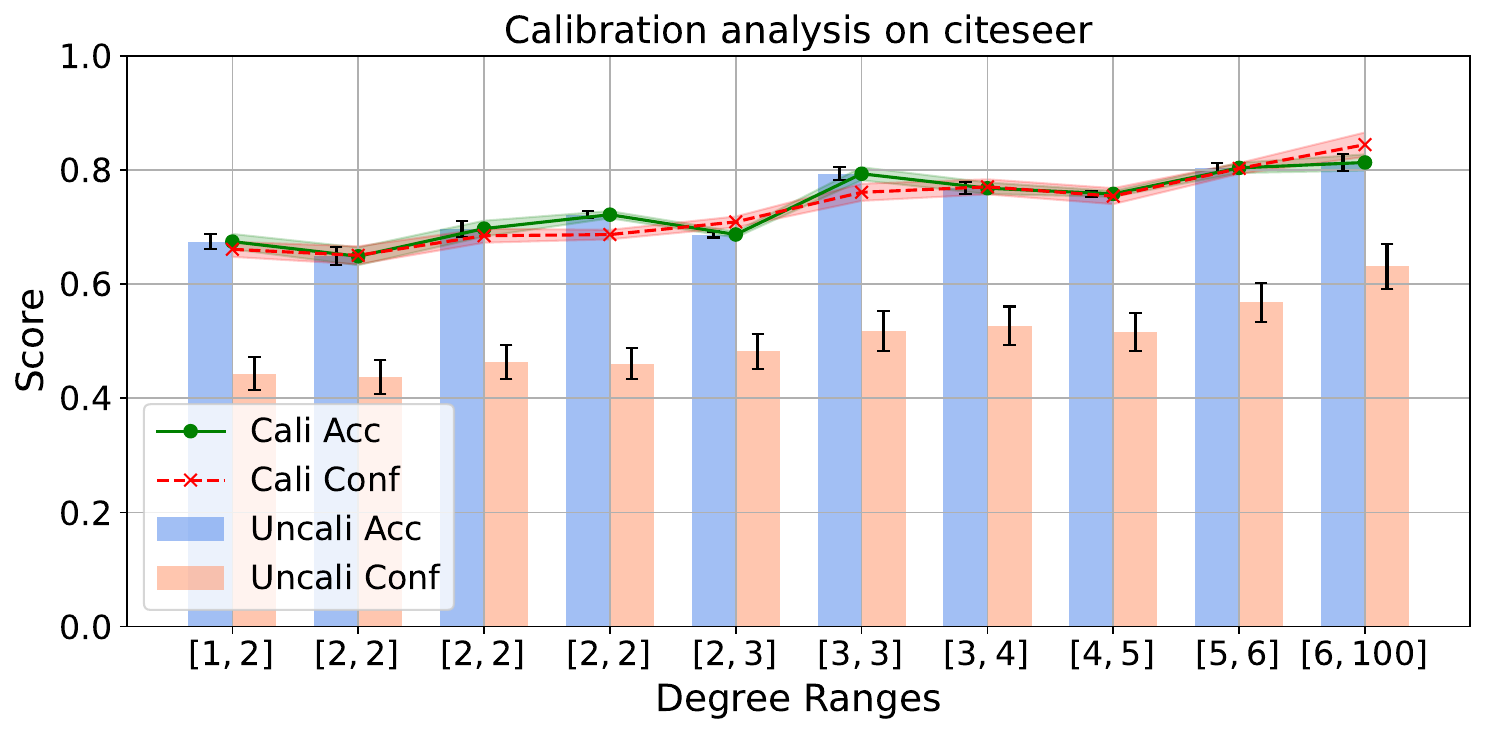}
        \caption{Degree-binned calibration analysis on citeseer.}
        \label{fig:citeseer_deg}
    \end{subfigure}
    \caption{In both plots, “Uncali” refers to the uncalibrated model and “Cali” refers to the calibrated model. (a) shows the reliability diagram comparing calibrated and uncalibrated outputs. The diagonal dashed line indicates perfect calibration (b) presents a degree-binned analysis of accuracy and confidence. Solid and dashed lines represent calibrated accuracy and confidence respectively. Results are averaged over 10 runs, with error bars indicating standard deviation. }

    \label{fig:citeseer_analysis}
\vspace{-5pt}
\end{figure}

To further illustrate this effect, we visualizes the calibration performance of WATS for Citeseer. The results in Figure ~\ref{fig:citeseer_analysis} shows a comprehensive analysis of calibration performance on the Citeseer dataset. In Figure ~\ref{fig:citeseer_reli}, the reliability diagram reveals that the uncalibrated model exhibits systematic underconfidence, with predicted probabilities consistently lower actual accuracy across bins. After calibration, the reliability curve aligns more closely with the diagonal, indicating a significant improvement in terms of confidence-accuracy alignment. Figure ~\ref{fig:citeseer_deg} complements this view by presenting a degree-binned analysis, with the degree increase we can observe a lower uncalibrated ECE, which also compile with our motivation. The uncalibrated model shows a clear mismatch between confidence and accuracy, especially for low-degree nodes, which tend to be the most under-confident. After applying calibration, confidence becomes well-aligned with accuracy across all degree ranges, with reduced standard deviation. This demonstrates that the method effectively improves overall calibration and enhances robustness, particularly for structurally sparse or uncertain regions. Full visualizations about the main experiment are provided in the Appendix.


Furthermore, GATS’s reliance on full attention over a node’s neighborhood leads to poor memory scalability and resulting in out-of-memory failures on large graphs such as Reddit. In contrast, WATS remains efficient by using spectral approximations to compute graph wavelet features, which improve the scalability of WATS.

\subsection{Ablation study}
\subsubsection{Different graph features}
To assess the effectiveness of graph wavelet features in post-hoc calibration, we conduct a comparative analysis against several widely used structural descriptors, including log-degree, betweenness centrality, clustering coefficient, and their various combinations, all evaluated under a consistent GCN-based framework. As summarized in Table~\ref{tab:ablation}, wavelet-based representations consistently yield superior calibration performance across most datasets. While certain individual features or their combinations may perform competitively on specific datasets, they tend to exhibit limited generalizability and often result in higher calibration error overall. This highlights the insufficiency of isolated structural indicators and underscores the necessity of incorporating rich, multiscale topological signals. In contrast, graph wavelet features demonstrate both effectiveness and robustness across diverse graph structures, suggesting that the information they encode captures nuanced patterns that cannot be fully replicated by aggregating conventional structural features.
\begin{table*}[ht]
\centering
\caption{Ablation study: ECE ($\downarrow$) comparison between graph wavelet and alternative structural features, where "Deg" denote log transformed degree, "Cen" denote betweenness centrality, "Clus" denote clustering coefficient, and ‘oom’ indicates out-of-memory failures where the method could not complete. Graph wavelet consistently outperforms other variants across most datasets.}
\label{tab:ablation}
\scriptsize
\setlength{\tabcolsep}{4pt}
\begin{tabular}{lcccccccc}
\toprule
\textbf{Dataset} & \textbf{Graph wavelet} & \textbf{Deg} & \textbf{Cen} & \textbf{Clus} & \textbf{Deg, Cen} & \textbf{Cen, Clus} & \textbf{Deg, Clus} & \textbf{Deg, Clus, Cen} \\
\midrule
Citeseer   & \textbf{2.15 $\pm$ 0.48} & 3.53 $\pm$ 1.16 & 3.10 $\pm$ 1.05 & 6.75 $\pm$ 1.44 & 7.24 $\pm$ 1.80 & 7.11 $\pm$ 1.80 & 7.10 $\pm$ 1.80 & 7.12 $\pm$ 1.69 \\
Computers  & \textbf{1.31 $\pm$ 0.29} & 1.61 $\pm$ 0.33 & 3.51 $\pm$ 0.83 & 2.75 $\pm$ 0.82 & 1.82 $\pm$ 0.22 & 2.78 $\pm$ 0.69 & 2.60 $\pm$ 0.71 & 2.72 $\pm$ 0.66 \\
Cora       & 2.13 $\pm$ 0.51 & 2.42 $\pm$ 0.72 & \textbf{1.86 $\pm$ 0.32} & 4.77 $\pm$ 0.40 & 4.43 $\pm$ 0.63 & 4.51 $\pm$ 0.48 & 4.51 $\pm$ 0.48 & 4.55 $\pm$ 0.40 \\
Cora-full  & \textbf{2.04 $\pm$ 0.13} & 3.08 $\pm$ 1.34 & 5.32 $\pm$ 0.18 & 5.66 $\pm$ 0.35 & 5.16 $\pm$ 0.23 & 5.19 $\pm$ 0.26 & 5.19 $\pm$ 0.26 & 5.18 $\pm$ 0.25 \\
Photo      & \textbf{1.03 $\pm$ 0.18}  & 1.32 $\pm$ 0.29 & 2.23 $\pm$ 0.40 & 1.85 $\pm$ 0.38 & 1.96 $\pm$ 0.30 & 1.93 $\pm$ 0.36 & 1.87 $\pm$ 0.34 & 1.89 $\pm$ 0.35 \\
Pubmed     & \textbf{1.04 $\pm$ 0.07} & 1.40 $\pm$ 0.35 & 2.90 $\pm$ 0.38 & 2.30 $\pm$ 0.29 & 1.83 $\pm$ 0.20 & 1.92 $\pm$ 0.22 & 1.93 $\pm$ 0.22 & 1.94 $\pm$ 0.20 \\
Reddit     & \textbf{1.02 $\pm$ 0.06} & 1.58 $\pm$ 0.17 & oom & oom & oom & oom & oom & oom \\
\bottomrule
\end{tabular}
\vspace{-10pt}
\end{table*}
\subsubsection{Sensitivity analysis of graph wavelet hyper-parameters.}
To assess the robustness of WATS, we perform an exhaustive grid search over the Chebyshev order
\(k \in \{2,3,4,5\}\) and the heat-kernel scale
\(s \in \{0.1,\,0.4,\,0.8,\,1.2,\,1.6,\,2.0,\,2.5\}\)
on seven node-classification benchmarks, we visualize the changes of ECE for varying $k$ and $s$ for Citeseer, Cora, Computers and Reddit on Figure ~\ref{fig:hype-analysis} (full results about hyperparameters are in Appendix).

\begin{figure}[h]
    \centering
    \begin{subfigure}[b]{0.245\textwidth}
        \centering
        \includegraphics[width=\textwidth]{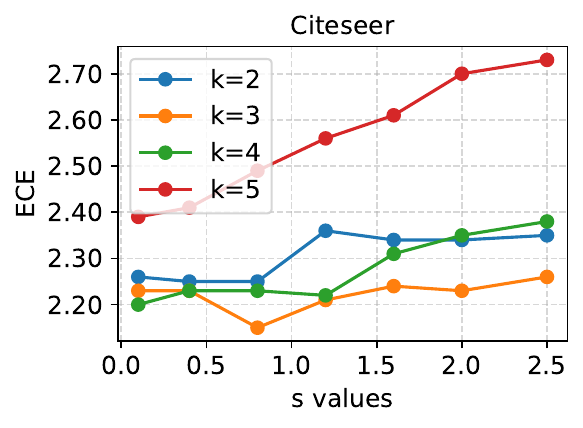}

        \label{fig:citeseer_line}
    \end{subfigure}
    \begin{subfigure}[b]{0.245\textwidth}
        \centering
        \includegraphics[width=\textwidth]{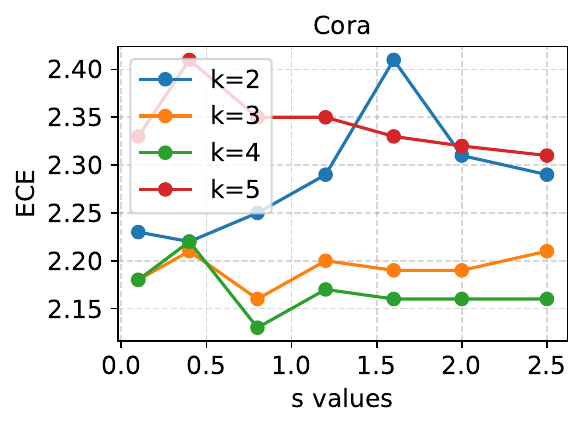}

        \label{fig:cora_line}
    \end{subfigure}
    \begin{subfigure}[b]{0.245\textwidth}
        \centering
        \includegraphics[width=\textwidth]{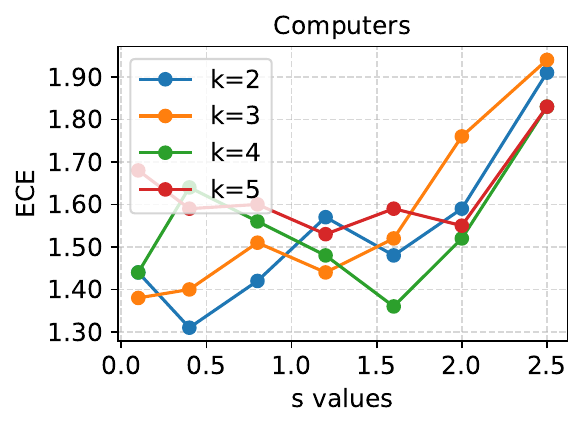}

        \label{fig:computers_line}
    \end{subfigure}
    \begin{subfigure}[b]{0.245\textwidth}
        \centering
        \includegraphics[width=\textwidth]{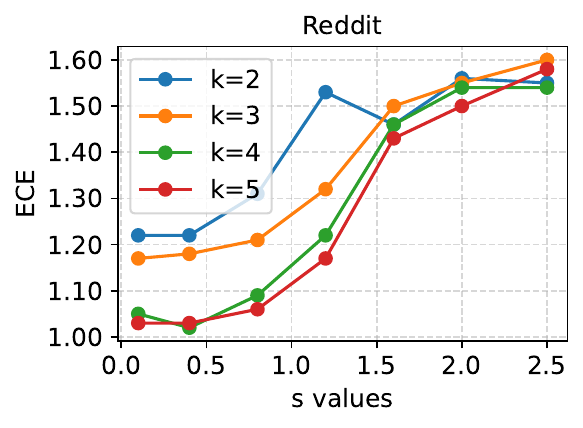}

        \label{fig:reddit_line}
    \end{subfigure}
    \caption{Sensitivity analysis of wavelet hyper-parameters. Each plot shows the ECE scores on different datasets with varying wavelet scale parameter $s$ (x-axis) and polynomial order $k$. Each line represents a different Chebyshev order $k$: blue for $k=2$, orange for $k=3$, green for $k=4$, and red for $k=5$.}
    \label{fig:hype-analysis}
\vspace{-5pt}
\end{figure}

The results demonstrate that calibration quality in heat-kernel graph wavelets is highly sensitive to both the Chebyshev order $k$ and the diffusion scale $s$. Across all evaluated benchmarks, the minimum ECE is consistently obtained when $k \in \{3,4\}$. Orders below this range restrict the receptive field to an overly local neighbourhood, omitting salient meso-scale structure, whereas excessively large orders indiscriminately propagate information from noisy nodes, thereby deteriorating calibration. We also observe that lower-connectivity graphs are generally more sensitive to $k$, as higher orders tend to introduce additional noise into the calibration process.

The diffusion scale $s$ exhibits a pronounced dependence on graph connectivity. In densely connected graphs, calibration is markedly sensitive to the choice of $s$; once $s$ surpasses its dataset-specific optimum, the ECE rises rapidly, indicating severe over-smoothing. By contrast, in sparser graphs, calibration performance is comparatively robust: deviations from the optimal $s$ induce only gradual changes in ECE, reflecting a more stable trade-off between local and global diffusion.

Importantly, although the result is highly correlated to hyperparameter choices, it remains consistently effective within a broad and practical range, specifically when $k \in \{3,4\}$ and $s \in [0.4,\,1.2]$. Even without fine-tuning, WATS typically outperforms existing calibration methods across this range, confirming its robustness and the strength of wavelet-based structural signals. These findings reinforce the hypothesis that effective calibration is primarily governed by localized structural cues, with densely connected graphs requiring more precise diffusion control.

\subsection{Complexity Analysis}

We further compare the complexity with other post-hoc calibration method to prove the computational efficiency. Our method consists of two main components: graph wavelet feature extraction and a two-layer MLP for temperature prediction. Let \( k \) be the Chebyshev polynomial order. Each Chebyshev term requires a sparse matrix multiplication, leading to a total time complexity of
\(
\mathcal{O}(k|\mathcal{E}| + |\mathcal{V}|k),
\)
where \( |\mathcal{E}| \) and \( |\mathcal{V}| \) denote the number of edges and nodes, respectively. The first term accounts for \( k \) sparse multiplications over the Laplacian, while the second accounts for the intermediate tensor concatenation and normalization steps. The wavelet features of each node (dimension \( k+1 \)) are passed through a two-layer MLP with hidden size \( h \). The per-node computation costs \( \mathcal{O}((k+1)h) \), and thus the total cost over all nodes is:
\(
\mathcal{O}(|\mathcal{V}|kh).
\)
Combining the above, the overall time complexity of our method is:
\[
\mathcal{O}(k|\mathcal{E}| + |\mathcal{V}|k + |\mathcal{V}|kh) = \mathcal{O}(k|\mathcal{E}| + |\mathcal{V}|kh).
\]
Compared to CaGCN \citep{wang2021confident} with \( \mathcal{O}(|\mathcal{E}|F + |\mathcal{V}|F^2) \), and GATS \citep{velivckovic2017graph} with \( \mathcal{O}(|\mathcal{E}|FH + |\mathcal{V}|F^2) \), our model is significantly more efficient, especially when \( F \) is large or multi-head attention is used. GETS \citep{zhuang2024gets} incurs higher cost due to expert selection, with complexity \( \mathcal{O}(k(|\mathcal{E}|F + |\mathcal{V}|F^2) + |\mathcal{V}|MF) \), where \( k \ll M \). In practice, the wavelet transformation can be precomputed and reused as a static input. 

\section{Conclusion}
We introduce Wavelet-Aware Temperature Scaling (WATS), a lightweight post-hoc calibration framework that assigns node-specific temperature values based on graph wavelet features. By exploiting localized structural representations, WATS effectively captures diverse structural characteristics and implicitly broadens each node's structural receptive field, thereby enhancing post-hoc information utilization with minimal computational overhead. Experimental results on seven benchmark datasets and two GNN backbones consistently show that WATS achieves the lowest ECE and markedly improves calibration stability, contributing to more reliable and trustworthy GNN predictions, especially for the high risk domains. While if the node distribution is extremely skewed, WATS may tend to be slightly overconfident for the large degree nodes, especially for the sparse graph due to the insufficient samples for calibration training.

Although WATS primarily captures rich topological information, future work could explore integrating structurally similar yet distant neighborhoods to introduce global structural context, which means capture the node with similar structural embedding. This could further enhance calibration performance and robustness, provided that the inclusion of such global information avoids introducing extraneous noise.

\clearpage
\bibliographystyle{plainnat} 
\bibliography{references}  

\clearpage
\appendix

\section{Experiment setting}
We randomly conduct the train test split 10 times for each dataset with identical random seed. We employed the GATS, GETS and CaGCN based on their paper and code.
The hyperparameters for backbone GNNs training are based on the complexity of graph data. The detail is given below Table ~\ref{tab:gnn_hyper}.

\begin{table}[h!]
    \centering
    \caption{Summary of training parameters of GCN and GAT}
    \label{tab:gnn_hyper}
    \begin{tabular}{lcccccc}
        \toprule
        \textbf{Dataset} & \textbf{Hidden Dim.} & \textbf{Dropout} &
        \textbf{Epochs} & \textbf{Learning Rate} & \textbf{Weight Decay} \\
        \midrule
        Citeseer   & 16  & 0.5 & 200 & $1\times10^{-2}$ & $5\times10^{-4}$ \\
        Computers  & 64  & 0.8 & 200 & $1\times10^{-2}$ & $1\times10^{-3}$ \\
        Cora-full  & 64  & 0.8 & 200 & $1\times10^{-2}$ & $1\times10^{-3}$ \\
        Cora       & 16  & 0.5 & 200 & $1\times10^{-2}$ & $5\times10^{-4}$ \\
        Photo      & 64  & 0.8 & 200 & $1\times10^{-2}$ & $1\times10^{-3}$ \\
        Pubmed     & 16  & 0.5 & 200 & $1\times10^{-2}$ & $5\times10^{-4}$ \\
        Reddit     & 16  & 0.5 & 200 & $1\times10^{-2}$ & $5\times10^{-4}$ \\
        \bottomrule
    \end{tabular}
\end{table}

Full details of WATS in calibration is given be on Table ~\ref{tab:hyper_wats}. Hidden dimension and drop out are chosen based on the data complexity. The hyperparameter of graph wavelet \(k\) and \(s\) are chosen based on the Ablation study.
\begin{table}[h!]
    \centering
    \caption{Calibration settings for WATS}
    \label{tab:hyper_wats}
    \begin{tabular}{lcccccc}
        \toprule
        \textbf{Dataset} & \textbf{Hidden Dim.} & \textbf{Dropout} &
        \textbf{k} & \textbf{s} \\
        \midrule
        Citeseer    & 32 & 0.4 & 3   & 0.8  \\
        Computers   & 64 & 0.4 & 2   & 0.4     \\
        Cora-full   & 128 & 0.2 & 4   & 1.2     \\
        Cora        & 16 & 0.95 & 4 & 0.8    \\
        Photo       & 32 & 0.4 & 4   & 0.4  \\
        Pubmed      & 32 & 0.4 & 4   & 1.6     \\
        Reddit      & 64 & 0.4 & 4  & 0.4    \\
        \bottomrule
    \end{tabular}
\end{table}

Full details of the Chosen datasets is given on Table ~\ref{tab:dataset_summary}. It reports the number of nodes, edges, average node degree, input feature dimensions, and number of classes for each dataset. These datasets cover a diverse range of graph sizes, densities, and classification tasks. This diversity ensures a comprehensive evaluation of the proposed method under varying structural and semantic conditions.
\begin{table}[h!]
\centering
\caption{Summary of selected datasets}
\label{tab:dataset_summary}
\begin{tabular}{lcccccc}
\toprule
\textbf{Dataset} & \textbf{\#Nodes} & \textbf{\#Edges} & \textbf{Avg. Degree} & \textbf{\#Features} & \textbf{\#Classes} \\
\midrule
Citeseer     & 3,327    & 12,431       & 7.4   & 3,703  & 6  \\
Computers    & 13,381   & 491,556      & 73.4  & 767    & 10 \\
Cora         & 2,708    & 13,264       & 9.7   & 1,433  & 7  \\
Cora-full    & 18,800   & 144,170      & 15.3  & 8,710  & 70 \\
Photo        & 7,487    & 238,087      & 63.6  & 745    & 8  \\
Pubmed       & 19,717   & 108,365      & 10.9  & 500    & 3  \\
Reddit       & 232,965  & 114,848,857  & 98.5  & 602    & 41 \\
\bottomrule
\end{tabular}
\end{table}

\paragraph{Computational Environment.}
All experiments are conducted using the following environment with PyTorch 2.4.0 (Python 3.11, CUDA 12.4.1), Hardware: NVIDIA GTX 4090 GPU with 32 GB RAM on Runpod cloud service (Ubuntu 22.04)

\section{Hyperparameter analysis results}
Here is the full result for the experiment on hyperparameter analysis. Tables ~\ref{tab:k2} to \ref{tab:k5} report the calibration performance measured by ECE of WATS under varying graph wavelet hyperparameters, specifically the Chebyshev order $k \in \{2, 3, 4, 5\}$ and diffusion scale $s \in \{0.1, 0.4, 0.8, 1.2, 1.6, 2.0, 2.5\}$. For each dataset, ECE values are presented across a range of $s$ values. 

\begin{table*}[h!]
\centering
\caption{ECE ($\downarrow$) for $k=2$ under varying diffusion scale $s$.}
\label{tab:k2}
\scriptsize
\begin{tabular}{lccccccc}
\toprule
\textbf{Dataset} & $s\!=\!0.1$ & $s\!=\!0.4$ & $s\!=\!0.8$ & $s\!=\!1.2$ & $s\!=\!1.6$ & $s\!=\!2.0$ & $s\!=\!2.5$\\
\midrule
Citeseer   & 2.26 $\pm$ 0.68 & 2.25 $\pm$ 0.66 & 2.25 $\pm$ 0.64 & 2.36 $\pm$ 0.72 & 2.34 $\pm$ 0.72 & 2.34 $\pm$ 0.72 & 2.35 $\pm$ 0.74 \\
Computers  & 1.44 $\pm$ 0.26 & 1.31 $\pm$ 0.29 & 1.42 $\pm$ 0.24 & 1.57 $\pm$ 0.35 & 1.48 $\pm$ 0.46 & 1.59 $\pm$ 0.43 & 1.91 $\pm$ 0.64 \\
Photo      & 1.31 $\pm$ 0.31 & 1.29 $\pm$ 0.24 & 1.59 $\pm$ 0.43 & 1.87 $\pm$ 0.34 & 1.99 $\pm$ 0.48 & 2.10 $\pm$ 0.48 & 2.27 $\pm$ 0.45 \\
Cora       & 2.23 $\pm$ 0.70 & 2.22 $\pm$ 0.65 & 2.25 $\pm$ 0.66 & 2.29 $\pm$ 0.67 & 2.41 $\pm$ 0.70 & 2.31 $\pm$ 0.69 & 2.29 $\pm$ 0.72 \\
Pubmed     & 1.13 $\pm$ 0.09 & 1.12 $\pm$ 0.10 & 1.09 $\pm$ 0.10 & 1.10 $\pm$ 0.10 & 1.05 $\pm$ 0.11 & 1.07 $\pm$ 0.13 & 1.20 $\pm$ 0.43 \\
Cora-full  & 2.82 $\pm$ 0.37 & 2.75 $\pm$ 0.36 & 2.76 $\pm$ 0.52 & 2.40 $\pm$ 0.66 & 2.68 $\pm$ 1.40 & 4.93 $\pm$ 1.06 & 5.27 $\pm$ 0.24 \\
Reddit     & 1.22 $\pm$ 0.08 & 1.22 $\pm$ 0.09 & 1.31 $\pm$ 0.19 & 1.53 $\pm$ 0.18 & 1.46 $\pm$ 0.16 & 1.56 $\pm$ 0.10 & 1.55 $\pm$ 0.09 \\
\bottomrule
\end{tabular}
\end{table*}

\begin{table*}[h!]
\centering
\caption{ECE ($\downarrow$) for $k=3$ under varying diffusion scale $s$.}
\label{tab:k3}
\scriptsize
\begin{tabular}{lccccccc}
\toprule
\textbf{Dataset} & $s\!=\!0.1$ & $s\!=\!0.4$ & $s\!=\!0.8$ & $s\!=\!1.2$ & $s\!=\!1.6$ & $s\!=\!2.0$ & $s\!=\!2.5$\\
\midrule
Citeseer   & 2.23 $\pm$ 0.50 & 2.23 $\pm$ 0.52 & 2.15 $\pm$ 0.48 & 2.21 $\pm$ 0.51 & 2.24 $\pm$ 0.52 & 2.23 $\pm$ 0.50 & 2.26 $\pm$ 0.49 \\
Computers  & 1.38 $\pm$ 0.32 & 1.40 $\pm$ 0.35 & 1.51 $\pm$ 0.20 & 1.44 $\pm$ 0.21 & 1.52 $\pm$ 0.28 & 1.76 $\pm$ 0.54 & 1.94 $\pm$ 0.75 \\
Photo      & 1.07 $\pm$ 0.18 & 1.22 $\pm$ 0.54 & 1.48 $\pm$ 0.58 & 1.63 $\pm$ 0.53 & 2.00 $\pm$ 0.46 & 2.11 $\pm$ 0.48 & 2.22 $\pm$ 0.46 \\
Cora       & 2.18 $\pm$ 0.44 & 2.21 $\pm$ 0.46 & 2.16 $\pm$ 0.46 & 2.20 $\pm$ 0.62 & 2.19 $\pm$ 0.62 & 2.19 $\pm$ 0.62 & 2.21 $\pm$ 0.61 \\
Pubmed     & 1.19 $\pm$ 0.09 & 1.14 $\pm$ 0.09 & 1.13 $\pm$ 0.13 & 1.10 $\pm$ 0.11 & 1.08 $\pm$ 0.10 & 1.18 $\pm$ 0.50 & 1.34 $\pm$ 0.62 \\
Cora-full  & 2.75 $\pm$ 0.21 & 2.82 $\pm$ 0.24 & 2.51 $\pm$ 0.37 & 2.32 $\pm$ 0.53 & 3.24 $\pm$ 1.58 & 4.81 $\pm$ 0.96 & 5.25 $\pm$ 0.29 \\
Reddit     & 1.17 $\pm$ 0.08 & 1.18 $\pm$ 0.07 & 1.21 $\pm$ 0.16 & 1.32 $\pm$ 0.19 & 1.50 $\pm$ 0.15 & 1.55 $\pm$ 0.07 & 1.60 $\pm$ 0.11 \\
\bottomrule
\end{tabular}
\end{table*}

\begin{table*}[h!]
\centering
\caption{ECE ($\downarrow$) for $k=4$ under varying diffusion scale $s$.}
\label{tab:k4}
\scriptsize
\begin{tabular}{lccccccc}
\toprule
\textbf{Dataset} & $s\!=\!0.1$ & $s\!=\!0.4$ & $s\!=\!0.8$ & $s\!=\!1.2$ & $s\!=\!1.6$ & $s\!=\!2.0$ & $s\!=\!2.5$\\
\midrule
Citeseer   & 2.39 $\pm$ 0.99 & 2.41 $\pm$ 0.62 & 2.49 $\pm$ 1.02 & 2.56 $\pm$ 0.97 & 2.61 $\pm$ 0.94 & 2.70 $\pm$ 0.99 & 2.73 $\pm$ 0.98 \\
Computers  & 1.68 $\pm$ 0.20 & 1.59 $\pm$ 0.27 & 1.60 $\pm$ 0.17 & 1.53 $\pm$ 0.22 & 1.59 $\pm$ 0.43 & 1.55 $\pm$ 0.29 & 1.83 $\pm$ 0.61 \\
Photo      & 1.06 $\pm$ 0.17 & 1.03 $\pm$ 0.18 & 1.39 $\pm$ 0.54 & 1.63 $\pm$ 0.57 & 1.86 $\pm$ 0.43 & 2.03 $\pm$ 0.45 & 2.12 $\pm$ 0.39 \\
Cora       & 2.18 $\pm$ 0.52 & 2.22 $\pm$ 0.52 & 2.13 $\pm$ 0.51 & 2.17 $\pm$ 0.53 & 2.16 $\pm$ 0.52 & 2.16 $\pm$ 0.51 & 2.16 $\pm$ 0.50 \\
Pubmed     & 1.19 $\pm$ 0.10 & 1.22 $\pm$ 0.10 & 1.16 $\pm$ 0.09 & 1.11 $\pm$ 0.08 & 1.04 $\pm$ 0.07 & 1.07 $\pm$ 0.09 & 1.06 $\pm$ 0.08 \\
Cora-full  & 2.53 $\pm$ 0.38 & 2.49 $\pm$ 0.32 & 2.32 $\pm$ 0.24 & 2.04 $\pm$ 0.24 & 3.26 $\pm$ 1.52 & 4.76 $\pm$ 1.07 & 5.19 $\pm$ 0.29 \\
Reddit     & 1.05 $\pm$ 0.06 & 1.02 $\pm$ 0.06 & 1.09 $\pm$ 0.23 & 1.22 $\pm$ 0.21 & 1.46 $\pm$ 0.13 & 1.54 $\pm$ 0.07 & 1.54 $\pm$ 0.08 \\
\bottomrule
\end{tabular}
\end{table*}

\begin{table*}[h!]
\centering
\caption{ECE ($\downarrow$) for $k=5$ under varying diffusion scale $s$.}
\label{tab:k5}
\scriptsize
\begin{tabular}{lccccccc}
\toprule
\textbf{Dataset} & $s\!=\!0.1$ & $s\!=\!0.4$ & $s\!=\!0.8$ & $s\!=\!1.2$ & $s\!=\!1.6$ & $s\!=\!2.0$ & $s\!=\!2.5$\\
\midrule
Citeseer   & 2.20 $\pm$ 0.49 & 2.23 $\pm$ 0.49 & 2.23 $\pm$ 0.49 & 2.22 $\pm$ 0.54 & 2.31 $\pm$ 0.46 & 2.35 $\pm$ 0.44 & 2.38 $\pm$ 0.45 \\
Computers  & 1.44 $\pm$ 0.32 & 1.64 $\pm$ 0.24 & 1.56 $\pm$ 0.20 & 1.48 $\pm$ 0.25 & 1.36 $\pm$ 0.27 & 1.52 $\pm$ 0.32 & 1.83 $\pm$ 0.54 \\
Photo      & 1.12 $\pm$ 0.42 & 1.38 $\pm$ 0.68 & 1.52 $\pm$ 0.75 & 1.78 $\pm$ 0.68 & 1.98 $\pm$ 0.51 & 2.09 $\pm$ 0.47 & 2.18 $\pm$ 0.45 \\
Cora       & 2.33 $\pm$ 0.72 & 2.41 $\pm$ 0.74 & 2.35 $\pm$ 0.76 & 2.35 $\pm$ 0.75 & 2.33 $\pm$ 0.75 & 2.32 $\pm$ 0.75 & 2.31 $\pm$ 0.75 \\
Pubmed     & 1.20 $\pm$ 0.11 & 1.23 $\pm$ 0.12 & 1.17 $\pm$ 0.08 & 1.17 $\pm$ 0.11 & 1.13 $\pm$ 0.10 & 1.09 $\pm$ 0.12 & 1.09 $\pm$ 0.12 \\
Cora-full  & 2.65 $\pm$ 0.90 & 2.34 $\pm$ 0.30 & 2.26 $\pm$ 0.23 & 1.90 $\pm$ 0.32 & 2.89 $\pm$ 1.53 & 4.09 $\pm$ 1.58 & 5.16 $\pm$ 0.31 \\
Reddit     & 1.03 $\pm$ 0.07 & 1.03 $\pm$ 0.07 & 1.06 $\pm$ 0.10 & 1.17 $\pm$ 0.15 & 1.43 $\pm$ 0.16 & 1.50 $\pm$ 0.09 & 1.58 $\pm$ 0.11 \\
\bottomrule
\end{tabular}
\end{table*}

\section{Full WATS visualizations}
\begin{figure}[H]
    \centering
    \begin{subfigure}[b]{0.45\textwidth}
        \centering
        \includegraphics[width=\textwidth, height=4cm]{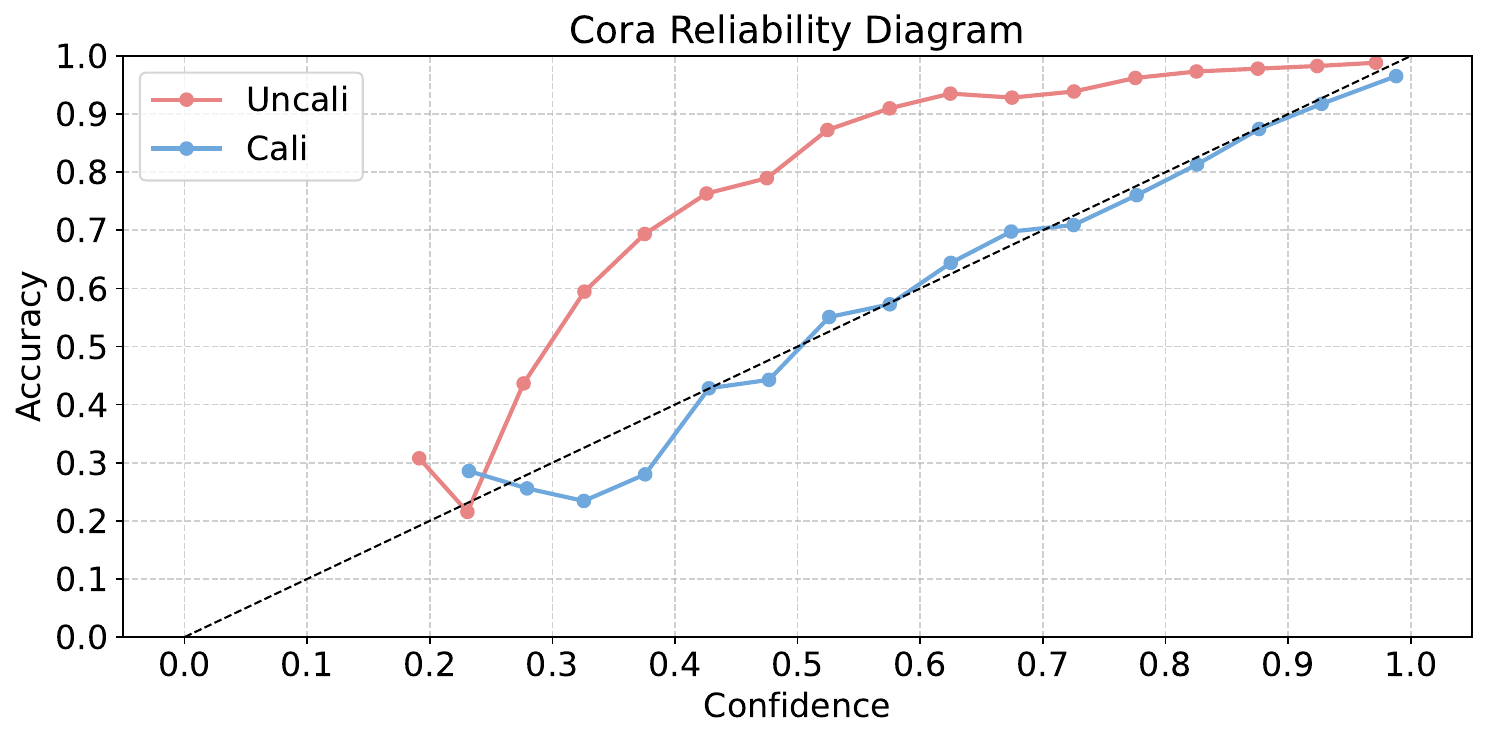}
        \caption{Reliability diagram on Cora.}
        \label{fig:cora_reli}
    \end{subfigure}
    \hfill
    \begin{subfigure}[b]{0.45\textwidth}
        \centering
        \includegraphics[width=\textwidth, height=4cm]{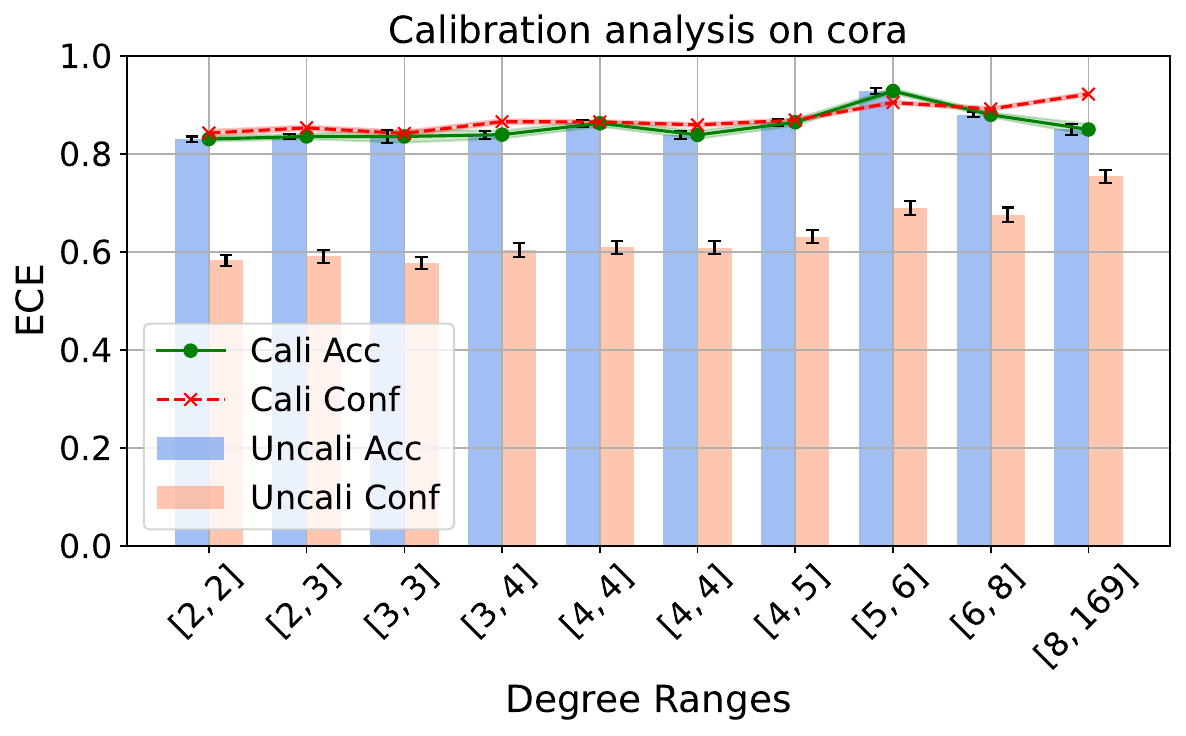}
        \caption{Degree-binned calibration analysis on Cora.}
        \label{fig:cora_deg}
    \end{subfigure}
    \caption{Calibration performance of Cora dataset.}
    \label{fig:cora_analysis}
\vspace{-10pt}
\end{figure}

\begin{figure}[H]
    \centering
    \begin{subfigure}[b]{0.48\textwidth}
        \centering
        \includegraphics[width=\textwidth, height=4cm]{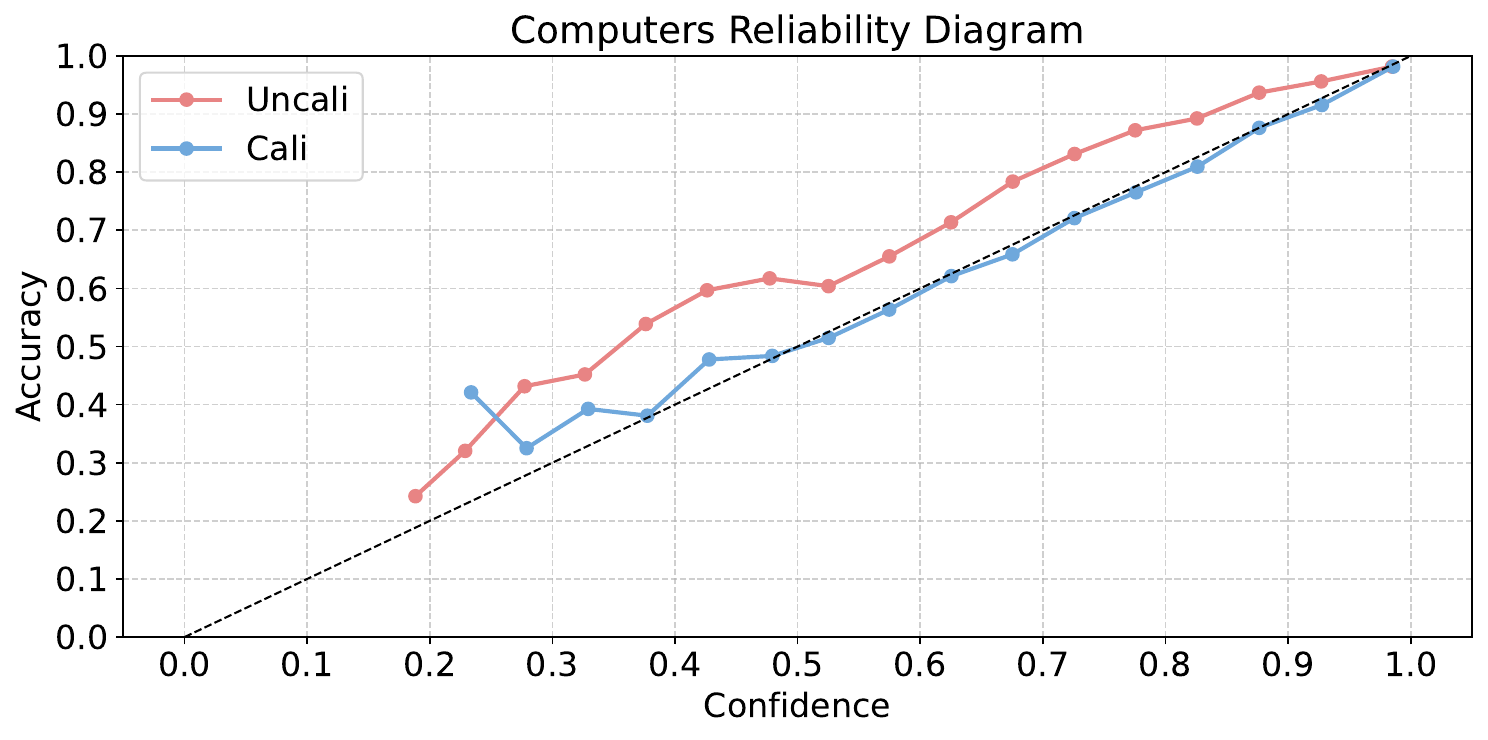}
        \caption{Reliability diagram on Computers.}
        \label{fig:computers_reli}
    \end{subfigure}
    \hfill
    \begin{subfigure}[b]{0.48\textwidth}
        \centering
        \includegraphics[width=\textwidth, height=4cm]{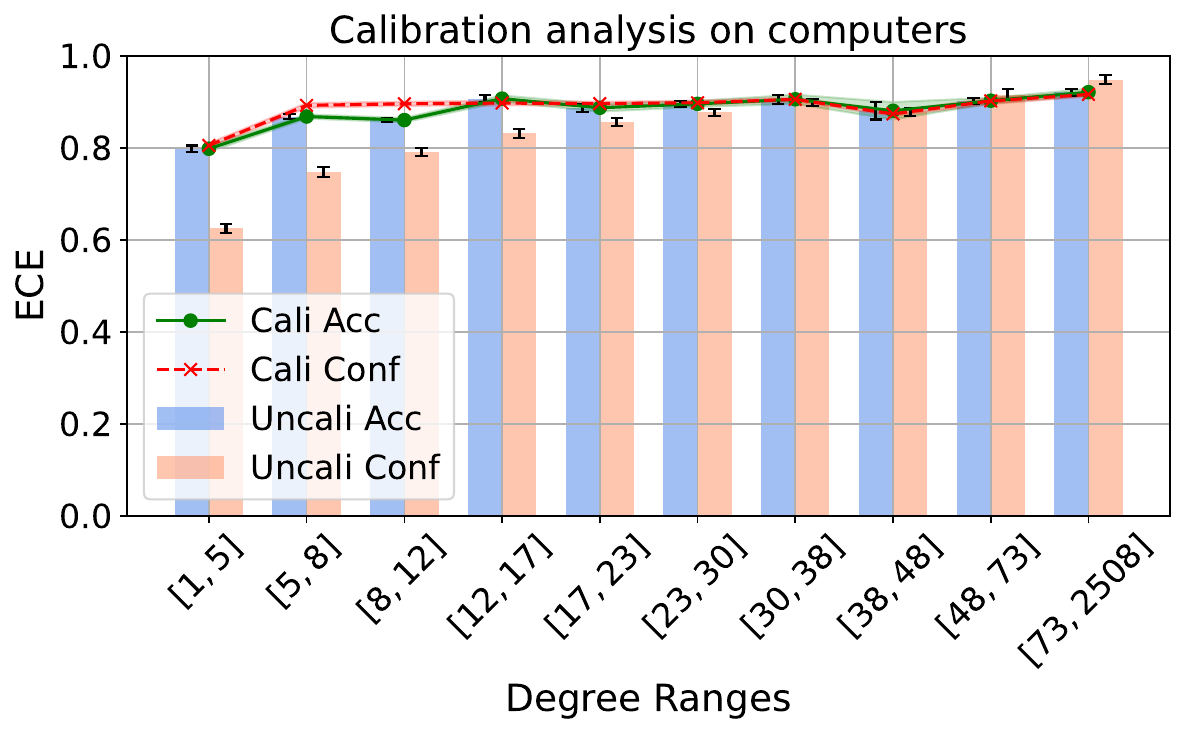}
        \caption{Degree-binned calibration analysis on Computers.}
        \label{fig:computers_deg}
    \end{subfigure}
    \caption{Calibration performance of Computers dataset.}
    \label{fig:computers_analysis}
\vspace{-7pt}
\end{figure}

\begin{figure}[H]
    \centering
    \begin{subfigure}[b]{0.45\textwidth}
        \centering
        \includegraphics[width=\textwidth, height=4cm]{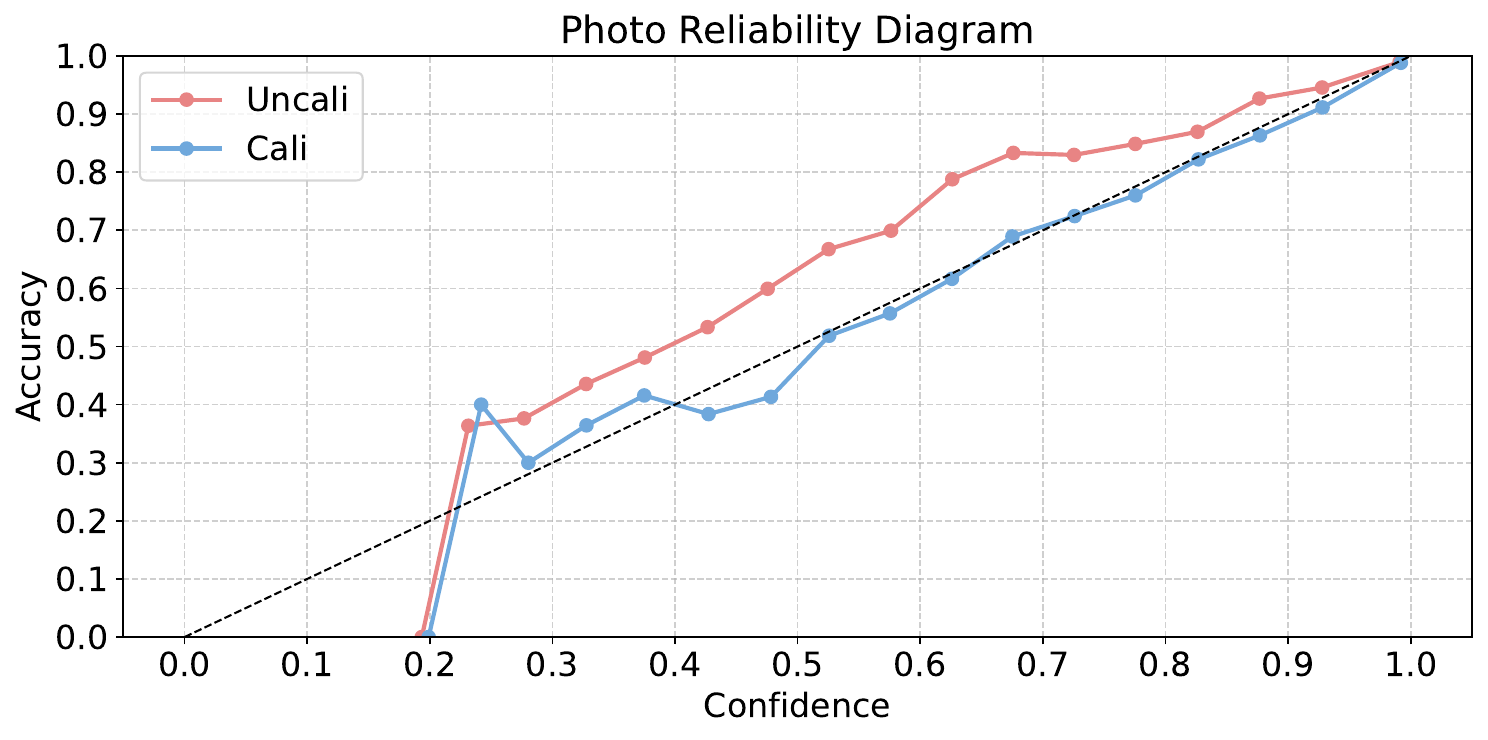}
        \caption{Reliability diagram on Photo.}
        \label{fig:photo_reli}
    \end{subfigure}
    \hfill
    \begin{subfigure}[b]{0.48\textwidth}
        \centering
        \includegraphics[width=\textwidth, height=4cm]{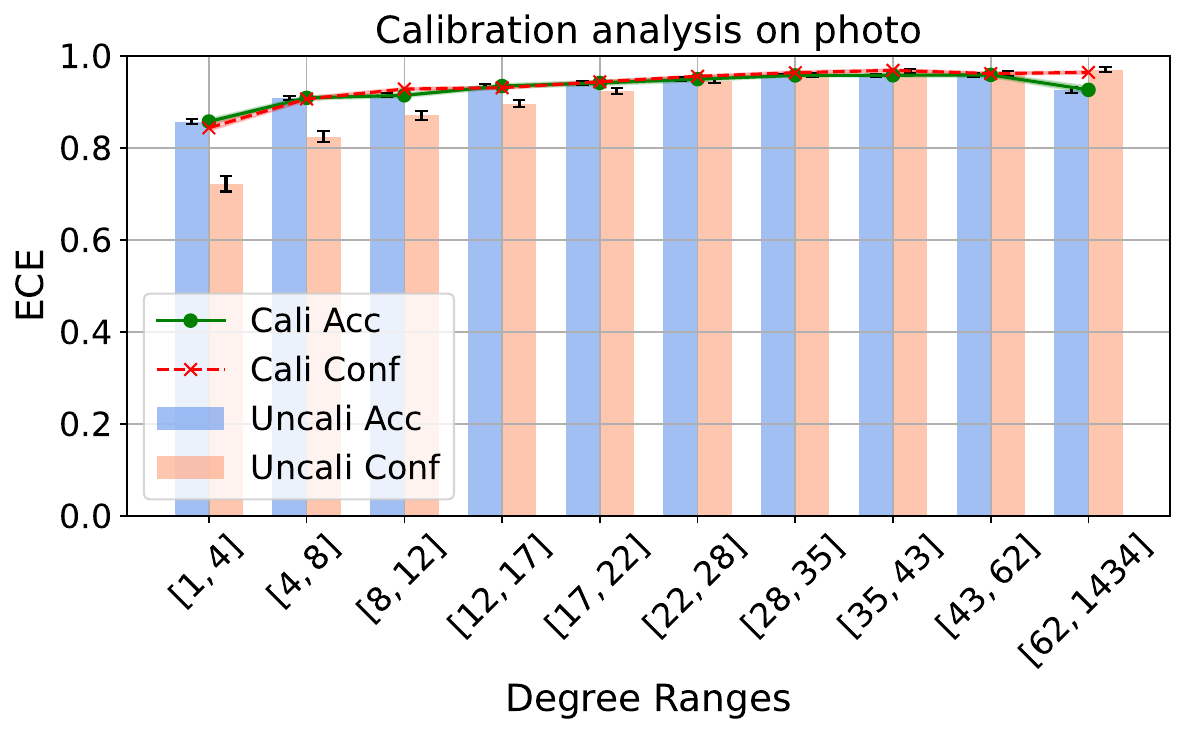}
        \caption{Degree-binned calibration analysis on Photo.}
        \label{fig:photo_deg}
    \end{subfigure}
    \caption{Calibration performance of Photo dataset.}
    \label{fig:photo_analysis}
\vspace{-7pt}
\end{figure}

\begin{figure}[H]
    \centering
    \begin{subfigure}[b]{0.45\textwidth}
        \centering
        \includegraphics[width=\textwidth, height=4cm]{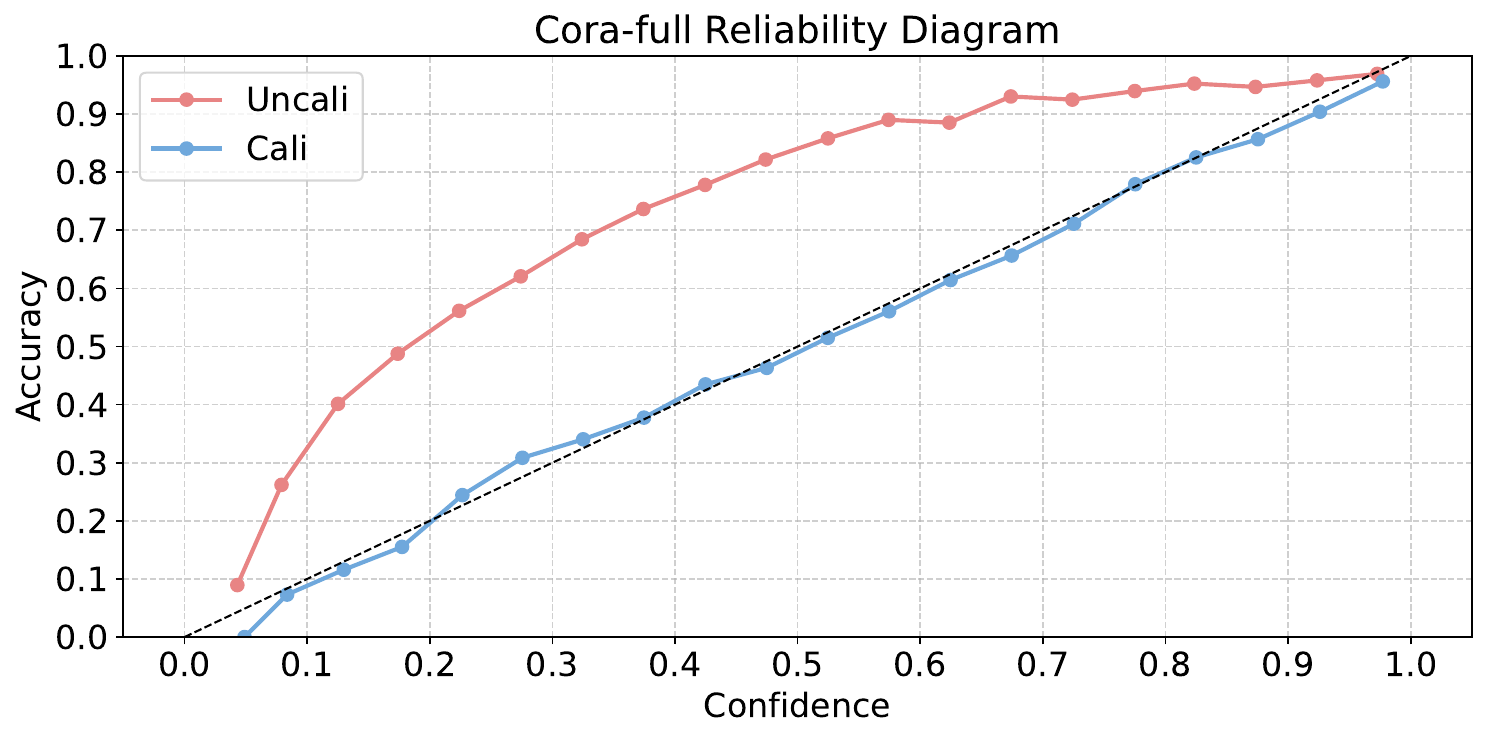}
        \caption{Reliability diagram on Cora-full.}
        \label{fig:corafull_reli}
    \end{subfigure}
    \hfill
    \begin{subfigure}[b]{0.48\textwidth}
        \centering
        \includegraphics[width=\textwidth, height=4cm]{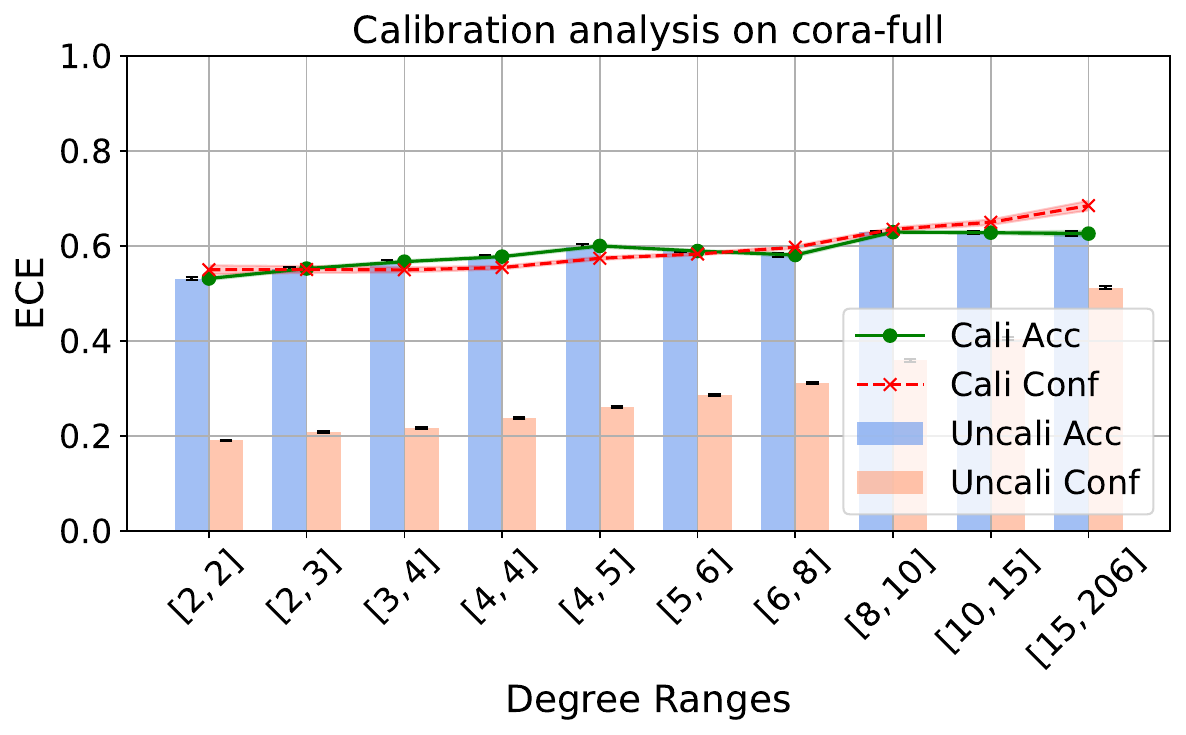}
        \caption{Degree-binned calibration analysis on Cora-full.}
        \label{fig:cora-full_deg}
    \end{subfigure}
    \caption{Calibration performance of Cora-full dataset.}
    \label{fig:cora-full_analysis}

\end{figure}

\begin{figure}[H]
    \centering
    \begin{subfigure}[b]{0.45\textwidth}
        \centering
        \includegraphics[width=\textwidth, height=4cm]{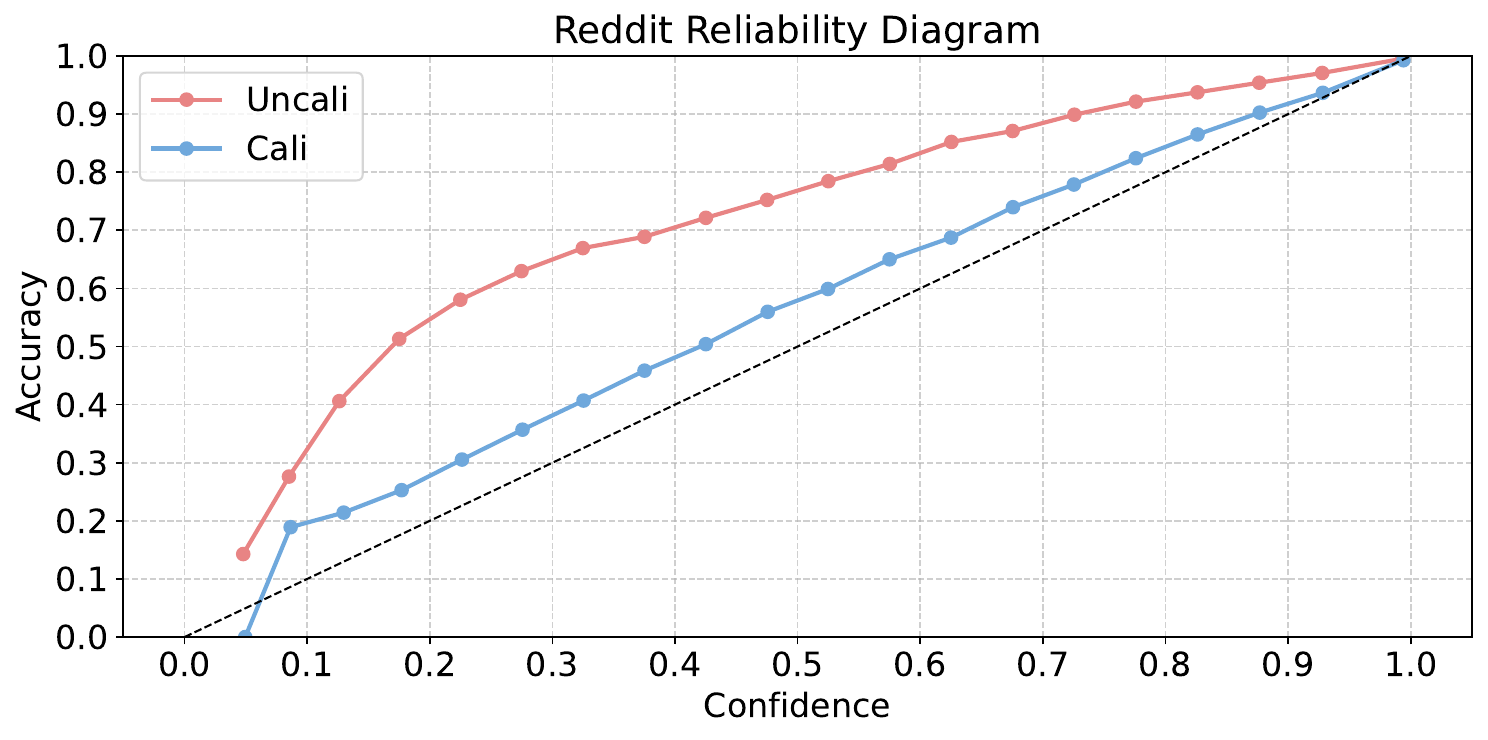}
        \caption{Reliability diagram on Reddit.}
        \label{fig:reddit_reli}
    \end{subfigure}
    \hfill
    \begin{subfigure}[b]{0.48\textwidth}
        \centering
        \includegraphics[width=\textwidth, height=4cm]{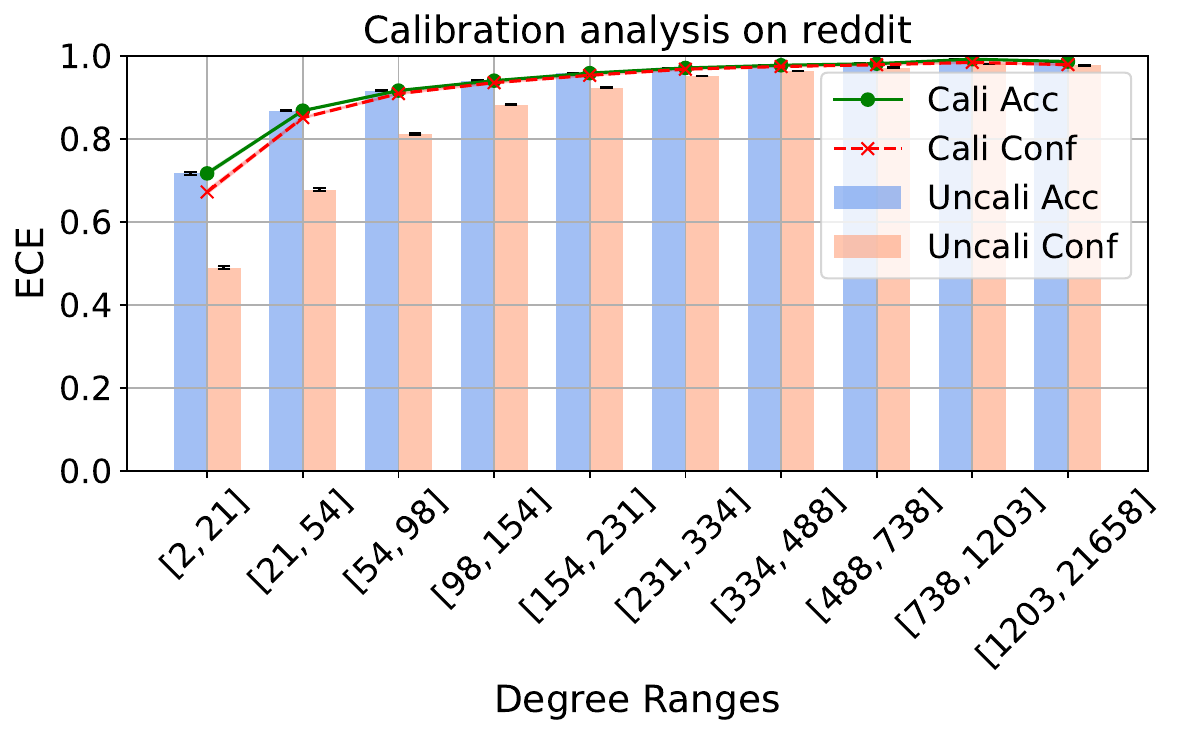}
        \caption{Degree-binned calibration analysis on Reddit.}
        \label{fig:reddit_deg}
    \end{subfigure}
    \caption{Calibration performance of Reddit dataset.}
    \label{fig:reddit_analysis}
\end{figure}

\begin{figure}[H]
    \centering
    \begin{subfigure}[b]{0.45\textwidth}
        \centering
        \includegraphics[width=\textwidth, height=4cm]{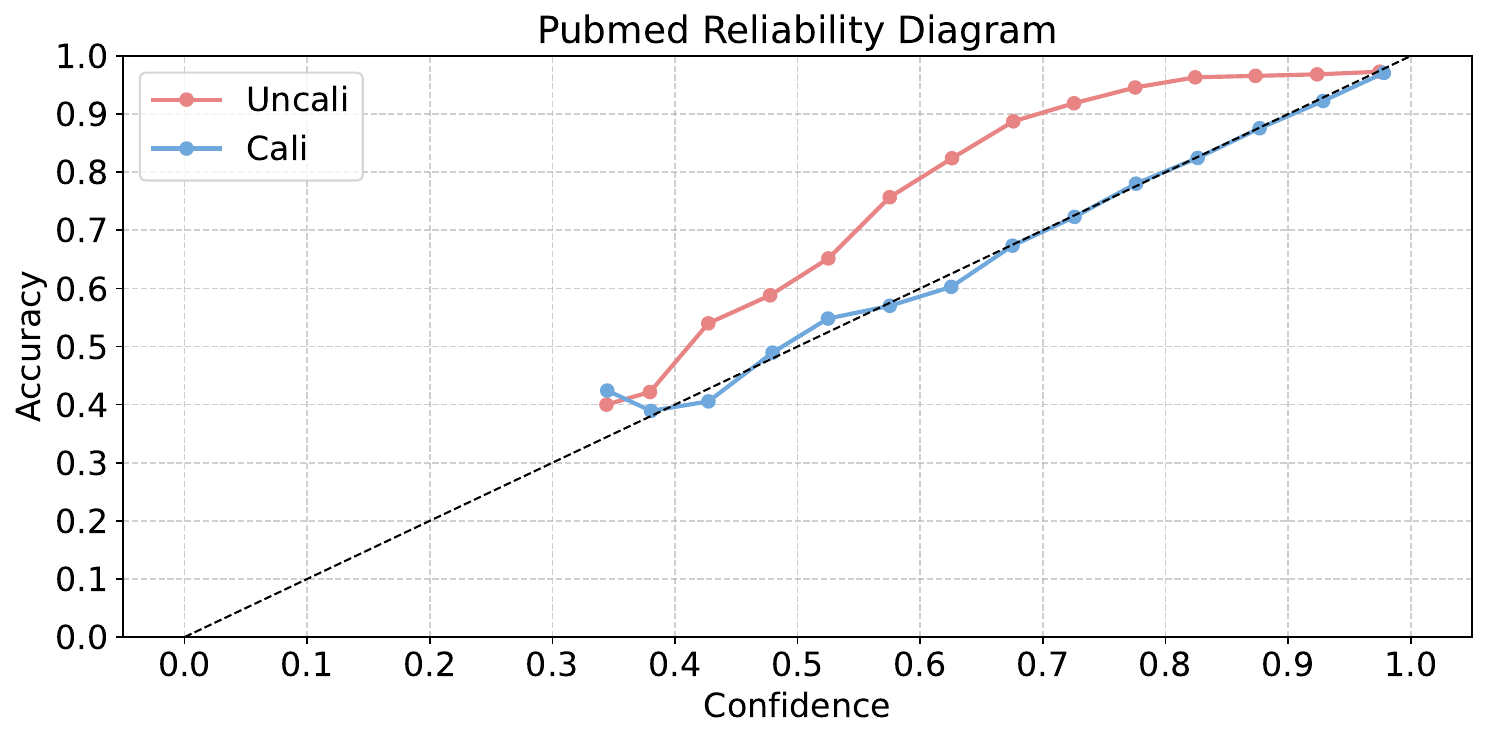}
        \caption{Reliability diagram on Pubmed.}
        \label{fig:pubmed_reli}
    \end{subfigure}
    \hfill
    \begin{subfigure}[b]{0.48\textwidth}
        \centering
        \includegraphics[width=\textwidth, height=4cm]{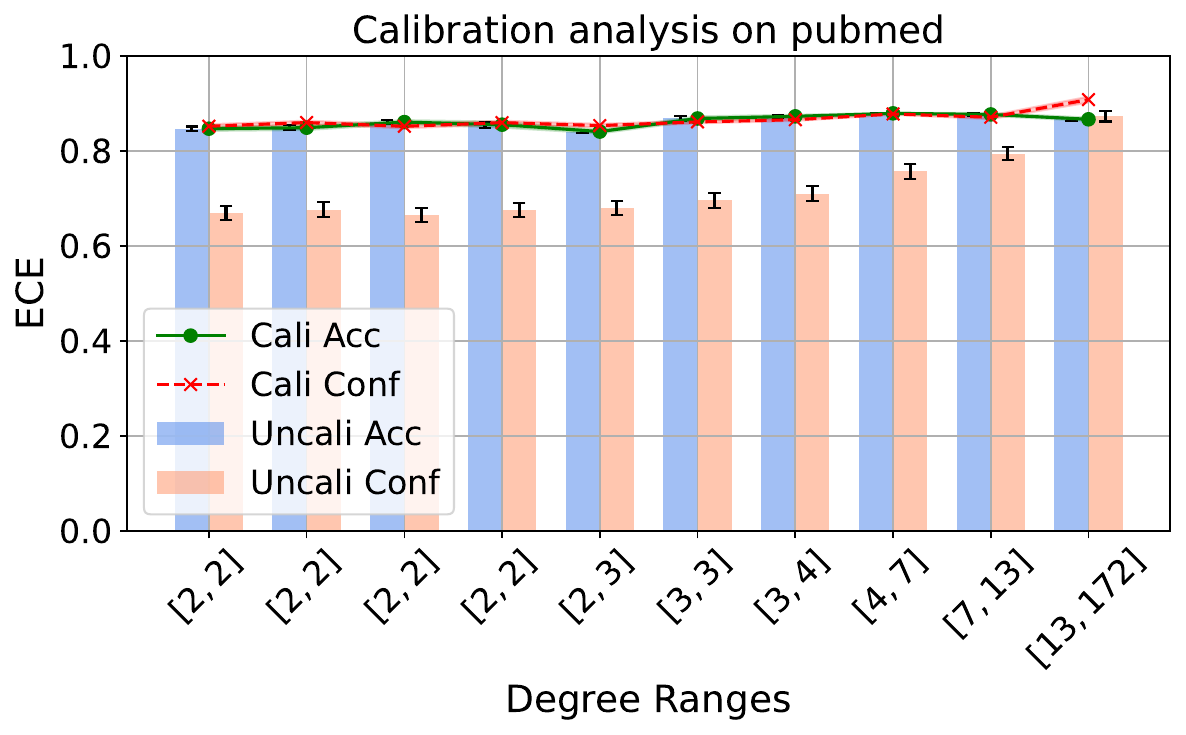}
        \caption{Degree-binned calibration analysis on Pubmed.}
        \label{fig:pubmed_deg}
    \end{subfigure}
    \caption{Calibration performance of Pubmed dataset.}
    \label{fig:pubmed_analysis}
\end{figure}

We provide the full visualizations of the calibration performance for WATS. Figures ~\ref{fig:cora_analysis} to ~\ref{fig:pubmed_analysis} illustrate the calibration performance of WATS on the other datasets. Each figure includes (a) a reliability diagram showing the alignment between predicted confidence and actual accuracy, and (b) a degree-binned analysis comparing confidence and accuracy before and after calibration. Results show that WATS significantly improves calibration and reduces the discrepancy between accuracy and confidence across all degree ranges and all confidence levels. Error bars indicate standard deviation over 10 runs.

\end{document}